\PassOptionsToPackage{hyperfootnotes=false}{hyperref}
\PassOptionsToPackage{sort&compress}{natbib}
\documentclass[times, review, 10pt]{elsarticle}
\journal{Pattern Recognition}

\usepackage{amsmath}   %
\usepackage{amssymb}   %
\usepackage{geometry}
\AddToHook{env/thebibliography/begin}{%
  \selectfont
  \setlength{\bibsep}{4pt plus 1pt}%
  \sloppy\hbadness=10000}

\usepackage{graphicx}
\usepackage{url}
\usepackage{float}
\usepackage{multirow}
\usepackage{threeparttable}
\usepackage{booktabs}
\usepackage{makecell}
\usepackage{bm}
\usepackage{enumitem}
\usepackage{xcolor}
\usepackage{colortbl}

\definecolor{first}{RGB}{0,153,0}
\definecolor{second}{RGB}{51,102,255}
\definecolor{third}{RGB}{204,102,0}

\usepackage{hyperref}
\AddToHook{env/tablenotes/begin}{\selectfont}
\hypersetup{
    colorlinks=true,
    linkcolor=blue,
    filecolor=magenta,
    urlcolor=cyan,
    citecolor=blue,
}

\usepackage{etoolbox}
\makeatletter
\patchcmd{\pprintMaketitle}{\Large\@title}
  {\fontsize{14}{17}\selectfont\@title}
  {}{\errmessage{PR port: title-size patch failed}}
\patchcmd{\pprintMaketitle}{\normalsize\elsauthors}
  {\footnotesize\elsauthors}
  {}{\errmessage{PR port: author-size patch failed}}
\makeatother

\begin{document}

\begin{frontmatter}

\title{Towards Automatic Video Annotation with ASH: Zero-Shot Open-Vocabulary Multi-Object Tracking and Segmentation}

\author[1]{Arash Rocky}
\ead{rocky@uwindsor.ca}

\author[1]{Q. M. Jonathan Wu\corref{cor1}}
\ead{jwu@uwindsor.ca}

\affiliation[1]{organization={Department of Electrical and Computer
Engineering, University of Windsor},
    city={Windsor},
    state={ON},
    postcode={N9B 3P4},
    country={Canada}}

\cortext[cor1]{Corresponding author.}

\begin{abstract}
Memory-attention-based Video Instance Segmentation (VIS) methods have
demonstrated strong zero-shot tracking capability, yet their substantial
memory requirements confine them to short video clips and their
single-prompt inference design makes multi-category open-vocabulary
tracking computationally prohibitive. This work introduces two
contributions toward fully automated tracking annotation of arbitrary
video. The \textbf{Generalized Presence Token (GPT)} reformulates SAM3's
inference pipeline to process $N$ text prompts simultaneously via virtual
prompt batching, reducing image encoding cost from $\mathcal{O}(N)$ to
$\mathcal{O}(1)$ with no modifications to any learned component. The
\textbf{Annotation and Segmentation Handler (ASH)} extends any
memory-attention VIS tracker to sequences of arbitrary length through
overlapping temporal chunks with IoU-based inter-chunk identity matching,
requiring no dataset-specific training. Instantiated on SAM3, the
resulting pipeline --- SAM3-ASH --- achieves state-of-the-art HOTA on
MOTS20 under fully zero-shot conditions and remains competitive with
trained specialists across seven additional benchmarks, while peak GPU
memory consumption stays below 25\,GB, establishing a practical baseline
for scalable, training-free automated video annotation.
\end{abstract}


\end{frontmatter}

\section{Introduction}
\label{S:1}

Recent advances in Vision-Language Models
(VLMs) have demonstrated
remarkable progress across image and video understanding tasks.
Despite these advances, the progress of VLMs in video understanding
remains bottlenecked by the scarcity and cost of large-scale annotated
video datasets --- a challenge that motivates the core contribution of
this work.

\begin{figure}[!t]
\centering
\begin{minipage}[t]{0.485\linewidth}\centering
\includegraphics[width=\linewidth]{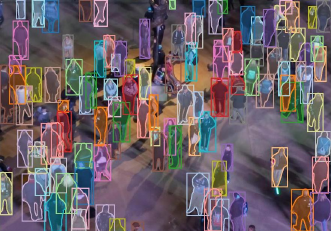}\\[1pt]
{\footnotesize $t=88$}
\end{minipage}\hfill
\begin{minipage}[t]{0.485\linewidth}\centering
\includegraphics[width=\linewidth]{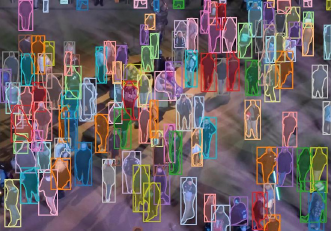}\\[1pt]
{\footnotesize $t=158$}
\end{minipage}
\caption{SAM3-ASH output on MOT20-04~\cite{MOT20} at $t{=}88$ and $t{=}158$
($2.8$\,s apart). Mask colours encode track identity: the orange- and
peach-outlined pedestrians at the right of each panel retain their identities
across the interval. Scenes of this density, with many small and mutually
occluding instances per frame, are precisely where manual annotation becomes
prohibitive. MOT20 is shown for illustration only; its box ground truth is not
commensurable with visible-mask output, so it is not used quantitatively.}
\label{fig1}
\end{figure}

\noindent\textbf{Dataset Annotation.}
Despite remarkable progress in computer vision, a critical bottleneck
remains unresolved: the creation of high-quality annotated
datasets. Manual annotation
is a time-consuming process that can extend over months, and is
inherently vulnerable to annotator bias and inconsistencies. The
financial burden of hiring expert annotators or utilizing data labeling
platforms can be substantial, particularly for specialized domains
requiring domain expertise. These challenges are further amplified when
dealing with complex scenarios, such as annotating small objects or
creating precise ground truth annotations for sequential frames, as
illustrated by the scene density in Fig.~\ref{fig1}. Moreover,
annotations are often tailored to specific methodological requirements,
forcing researchers to repeatedly re-annotate datasets to align with
their specific research needs --- a redundant and resource-intensive
process that impedes research progress.

\noindent\textbf{Annotation Scope.}
In this work, annotation refers to object tracking annotation
specifically: assigning consistent identities to object instances
across video frames, with bounding boxes derived from segmentation
masks as the output format. We argue that
automating this foundational annotation type can pave the way for
generating richer annotation types --- segmentation masks, action
labels, event detections --- with minimal additional effort.

\noindent\textbf{Existing Approaches.}
Prior work has explored self-supervised learning, data augmentation,
and annotation-free methods to reduce labeling cost, but none address the specific challenge of
consistent multi-object identity tracking across unconstrained video
sequences.

\noindent\textbf{Proposed Solution.}
The field still lacks a robust automatic annotation method capable of
generating high-quality datasets with minimal human intervention.
We propose that automated tracking annotation can be achieved by
building upon recent advances in open-vocabulary object
detection~\cite{GroundingDINO, GLEE}. However, the
central challenge remains maintaining consistent object identities
across frames --- motivating the need for reliable long-term tracking.
And this is where memory-attention-based video segmentation can fill
this gap. Toward this goal, recent work on SAM2~\cite{SAM2} advanced
automated video annotation by introducing video tracking capabilities
that allow human annotators to provide first-frame annotations and
refine subsequent frames using positive and negative clicks. Although
this reduces the manual effort, human intervention remains necessary.
SAM3~\cite{SAM3} took a further step toward a fully automated
annotation engine by integrating an open-vocabulary object detector
and a masklet matcher into SAM2. The detector employs a
noun-phrase-based prompting strategy and a learned presence token that
disentangles object existence from localization, reducing false
positives. The matcher further consolidates redundant segmentation
masks across frames. Despite these advances, SAM3's design introduces
substantial memory overhead on top of the already computationally
demanding SAM2 backbone, constraining its application to short video
clips on high-end hardware and limiting inference to a single prompt
per forward pass, making multi-category open-vocabulary tracking
computationally prohibitive.

Our response to these two constraints is deliberately training-free,
and its conceptual core is architectural rather than compositional.
We observe that SAM3's presence token --- introduced as a single-prompt
construct --- is implicitly prompt-indexed: by reformulating the
inference batch as a virtual batch of $N$ text prompts over a shared
visual encoding, the decoder can predict presence, localization, and
segmentation for $N$ independent concepts in a single forward pass
without modifying any learned component. This Generalized Presence
Token (GPT) collapses the image-encoding cost of open-vocabulary
multi-object inference from $\mathcal{O}(N)$ to $\mathcal{O}(1)$,
turning multi-category tracking from prohibitive to practical.
The memory constraint is addressed by the Annotation and Segmentation
Handler (ASH), a model-agnostic chunked inference layer compatible with
any memory-attention Video Instance Segmentation (VIS) method --- a
claim we substantiate empirically with two instantiations spanning two
backbones and two prompt modalities: box-prompted SAM2 (as FLASH,
prototyped in our earlier SAM2Auto pipeline~\cite{SAM2Auto}) and
text-prompted SAM3 (Appendix~\ref{sub:ash_sam2}).
The resulting pipeline, SAM3-ASH, serves as a strong baseline for
automated tracking annotation, demonstrating the practical viability of
memory-attention-based VIS on arbitrary-length video sequences.
SAM3-ASH minimizes the need for manual annotation, ensures consistent
object identification across frames, and operates without
dataset-specific training or extensive parameter tuning.

\noindent\textbf{Why Training-Free.}
Training-free operation is not an implementation convenience but a
requirement of the annotation setting itself. An annotation engine that
must be trained is circular: it presupposes labels in the very domain it
is meant to label. Keeping every learned component frozen removes this
circularity --- SAM3-ASH requires no seed annotations, no per-domain
retraining, and extends to novel vocabularies at deployment time by
editing a text prompt. The gaps to trained specialists reported in
Section~\ref{S:4} are the price of this property; the results show the
price is consistently small and, on several benchmarks, nonexistent.

\noindent\textbf{Contributions.}
The main contributions of this paper are as follows:
\begin{enumerate}
    \item \textbf{Generalized Presence Token}: The presence token in
    SAM3 was implicitly a single-prompt construct. We generalize it to
    a multi-presence token by reformulating the inference batch as a
    prompt-indexed virtual batch over a shared visual representation.
    This allows the decoder to simultaneously predict object presence,
    localization, and segmentation for $N$ independent semantic concepts
    in a single forward pass, reducing redundant image encoding from
    $\mathcal{O}(N)$ to $\mathcal{O}(1)$.

    \item \textbf{A}nnotation and \textbf{S}egmentation \textbf{H}andler
    \textbf{(ASH)}: A model-agnostic Multi-Object Tracker (MOT) that
    extends any memory-attention-based Video Instance Segmentation (VIS)
    model to process video sequences of arbitrary length. ASH overcomes
    the inherent memory constraints of VIS models by decomposing long
    sequences into temporal chunks while maintaining consistent object
    identities across boundaries, enabling fully automated video
    annotation without dataset-specific training or human intervention.

    \item We integrate the Generalized Presence Token and ASH into
    SAM3, forming SAM3-ASH, and demonstrate through extensive
    experiments across eight benchmarks that SAM3-ASH achieves
    competitive zero-shot tracking performance without dataset-specific
    training, establishing a practical baseline for fully automated
    tracking annotation of arbitrary video sequences. We further
    validate ASH's model-agnostic design empirically by comparing
    against a second instantiation on box-prompted
    SAM2~\cite{SAM2Auto}.
\end{enumerate}

The remainder of this paper is organized as follows:
Section~\ref{S:2} reviews related work in the areas closest to
automated annotation. Section~\ref{S:3} details the Generalized
Presence Token and the Annotation and Segmentation Handler (ASH) that
together form SAM3-ASH. Section~\ref{S:4} presents experimental
results and discussion. Section~\ref{S:5} outlines future directions,
and Section~\ref{S:6} concludes the paper.

\section{Related Works}
\label{S:2}

\subsection{Multi-Prompt Grounding in Vision-Language Models}
\label{MultiPrompt}
Existing Vision-Language Models (VLMs) address multi-concept grounding through four main strategies. Grounding DINO~\cite{GroundingDINO} concatenates all category names into a single text string and applies block-diagonal attention masking to isolate sub-phrases, coupling all concepts within one encoder context and risking concept interference as the prompt count grows. 
Autoregressive Large Vision-Language Models (LVLMs) such as Qwen3-VL~\cite{Qwen3-VL} generate bounding box coordinates token-by-token from the full prompt sequence, serializing multi-concept grounding through the decoder and providing no explicit binary presence signal per concept. 
Furthermore, Perception Encoder~\cite{PerceptionEncoder} takes a deep-fusion approach, processing the prompt list into discrete concept tokens that recontextualize image features inside the encoder before detection, though this requires retraining for new prompt distributions. 
In contrast, our Generalized Presence Token reformulates inference batching so that $N$ prompts are processed as parallel slots over a single cached image encoding, generalizing the presence token from a scalar to a prompt-indexed vector $\mathbf{P} \in \mathbb{R}^{L \times N \times Q}$ and achieving concept-decoupled presence scores and segmentation masks in one forward pass with $O(1)$ image encoding cost and no modifications to any learned component.

\subsection{Robust Tracking for Auto Annotation}
\label{sub:RobTrk}
Tracking methods in video understanding can be broadly categorized into three groups: Multiple Object Tracking (MOT) trackers operating on bounding boxes \cite{TrackFormer, DiffMOT, GHOSTTracker, Samba, Sparsetrack, SUSHI}; Multi-Object Tracking and Segmentation (MOTS) trackers \cite{masa} as an extension of MOT to pixel-level masks; and Video Object Segmentation (VOS), Interactive VOS (iVOS), and Video Instance Segmentation (VIS)~\cite{OVIS, PRVISsurvey} as mask-based instance-level approaches.

Within the MOT group, recent work refines the association stage itself: MSPNet~\cite{ZHENG2024110369} couples motion-guided feature aggregation with an occlusion-aware hierarchical matcher, while C-TWiX~\cite{CTWiX} replaces IoU at inference with a Transformer that learns coordinate-only tracklet affinities. ASH likewise treats association as a modular layer, but keeps it training-free so it remains agnostic to the upstream segmentation backbone.

Despite the lack of large-scale annotated datasets in each category \cite{TrackFormer, Samba, SAM2}, tracking methods have remained largely independent across categories, with little convergence between their approaches. As a solution, OmniTracker \cite{OmniTracker} introduced a \textbf{unified tracking-with-detection} architecture to integrate bounding box trackers (MOT) with mask trackers (VOS, iVOS, VIS). However, this method requires dataset-specific training, making it impractical for automated tracking annotation.

To address this limitation, SPAM~\cite{spam2024eccv} was developed for auto tracking annotation, leveraging synthetic pretraining, pseudo-labeling, and active learning with a graph-based model. While it focuses \textbf{primarily on MOT annotation} and achieves strong results on bounding-box benchmarks such as MOT17 and DanceTrack, it remains a closed-vocabulary specialist requiring additional training, and its bounding-box-annotated evaluation setting does not expose the ground truth misalignment that affects segment-based trackers on the same sequences.

\textbf{For VOS and iVOS}, SAM2 \cite{SAM2} introduced a memory bank-based approach for efficient mask tracking. Later, SAM3 \cite{SAM3} extended this with prompt-based grounding and a mask matcher to reduce redundant segments, taking a further step toward automated tracking annotation. While both models can track objects with well-defined boundaries, including subparts, they struggle with long video sequences and large numbers of objects due to memory constraints.

SAM2MOT~\cite{Sam2mot} further extends SAM2 to multi-object tracking by introducing a trajectory manager and a cross-object interaction module that detects and purges corrupted memory entries under occlusion, establishing the practical upper bound for memory-attention segmentation models on DanceTrack. However, its occlusion handling is designed for pairwise near-uniform-appearance scenarios and does not generalize to multi-category or long-sequence settings.

Concurrently, WorldTrace~\cite{WorldTrace} pursues bounded visual
memory on the generation side: a training-free framework that
compresses the growing key--value cache of video world models into a
fixed, addressable budget. Where WorldTrace restructures memory
\emph{inside} the model, ASH bounds it \emph{externally} through
model-agnostic temporal chunking, leaving the base model's memory
mechanism untouched.

We introduce \textbf{SAM3-ASH}, a unified bounding box (MOT) and mask-based (VOS, iVOS, VIS) tracker for annotation that extends SAM3's trackability while eliminating its memory constraints. Unlike existing memory bank-based trackers, SAM3-ASH addresses both the prompt-count bottleneck of open-vocabulary inference --- through GPT --- and the sequence-length constraint of memory-attention models --- through ASH --- requiring no additional training beyond the pretrained SAM3 weights.

\section{Auto Tracking Annotation Method}
\label{S:3}
In this section, we introduce SAM3-ASH, a fully automated video tracking annotation pipeline built upon two core contributions: the Generalized Presence Token (GPT), which extends SAM3's single-prompt grounding mechanism to simultaneously process multiple semantic concepts in a single forward pass, and the Annotation and Segmentation Handler (ASH), a model-agnostic multi-object tracker that extends any memory-attention-based VIS model to annotate video sequences of arbitrary length. Together, these components enable high-quality instance-level tracking annotations across large-scale datasets without dataset-specific training or human intervention. The following subsections detail each contribution and its role in the overall pipeline.

\subsection{Generalized Presence Token (GPT)}

SAM3 processes one semantic prompt per forward pass. Grounding $N$ concepts
therefore requires $N$ full encoder evaluations---an $O(N)$ cost that scales
poorly at inference time and prevents simultaneous multi-concept reasoning.
We address both limitations with the \textbf{Generalized Presence Token (GPT)},
which extends SAM3's single-prompt presence mechanism to $N$ independent
presence decisions computed in one pass, with image encoding cost reduced
to $O(1)$.

\paragraph{Virtual Prompt Batching.}
The visual backbone produces prompt-agnostic features. For image $I$, the
encoded representation $\mathbf{F} = f_\theta(I)$ is identical regardless of
which concept is queried, making repeated encoder calls redundant. We exploit
this by constructing a \emph{virtual batch} of size $N$ over a single cached
image feature. Given $N$ text prompts $\{\tau_i\}_{i=1}^{N}$, we define index
vectors:
\begin{equation}
    \mathbf{u}^{\text{img}} = \bigl[\,\underbrace{0,\,0,\,\ldots,\,0}_{N}\,\bigr]^\top,
    \qquad
    \mathbf{u}^{\text{txt}} = \bigl[0,\,1,\,\ldots,\,N{-}1\bigr]^\top,
\end{equation}
where $\mathbf{u}^{\text{img}}$ routes every batch slot to the single shared
visual feature, and $\mathbf{u}^{\text{txt}}$ routes each slot to its distinct
text embedding $\mathbf{t}_i$. The encoder and decoder run once at effective
batch size $N$, with cross-attention aligning slot $i$ to $\mathbf{t}_i$ while
drawing from the shared $\mathbf{F}$. The cost of adding a prompt is thus
reduced to text encoding and one decoder pass---the dominant backbone cost does
not scale with $N$.

\paragraph{Generalized Presence Decoding.}
In SAM3, the decoder emits a presence tensor
$\mathbf{p} \in \mathbb{R}^{L \times 1 \times Q}$, where $L$ is the number of
decoder layers and $Q$ is the number of object queries. Under virtual batching,
this tensor naturally generalizes to
$\mathbf{P} \in \mathbb{R}^{L \times N \times Q}$,
yielding an independent presence score $p_{i,q}^{(l)}$ for every
(layer~$l$, prompt~$i$, query~$q$) triple without any modification to the
decoder itself. The joint detection score is:
\begin{equation}
    s_{i,q} \;=\; \sigma\!\left(\hat{s}_{i,q}\right)
               \cdot \sigma\!\left(p_{i,q}\right),
\end{equation}
where $\hat{s}_{i,q}$ is the dot-product class logit between the $q$-th object
query and the $i$-th text embedding, and $\sigma$ denotes the sigmoid function.
Each prompt $\tau_i$ receives its own ranked set of detections, fully decoupled
from the remaining $N{-}1$ prompts.

\paragraph{Extension to Video.}
We carry virtual batching through the temporal dimension. At frame $t$, we set
$\mathbf{u}^{\text{img}}_t = [t,\,t,\,\ldots,\,t]^\top$ and reuse
$\mathbf{u}^{\text{txt}} = [0,\,1,\,\ldots,\,N{-}1]^\top$ across all frames.
Geometric prompt buffers and empty prompt constants are resized from batch
size~$1$ to~$N$; all other pipeline components---memory encoding, propagation,
and hotstart logic---remain unchanged. The result is a full $N$-concept
segmentation mask per frame from a single detector call, without any
per-prompt re-encoding.

\paragraph{Zero-Weight Generalization.}
Our contribution requires no changes to the transformer encoder, decoder,
segmentation head, or presence scoring head. Generalization is confined to the
construction of $\mathbf{u}^{\text{img}}$ and $\mathbf{u}^{\text{txt}}$ at the
inference batching layer. This confirms that the presence token in SAM3 is
architecturally multi-prompt capable; the original single-query design is an
interface constraint, not a model limitation. GPT unlocks parallel multi-concept
grounding as a zero-weight-change inference extension.

\begin{figure}[!t]
\centering
\includegraphics[trim={0.5cm, 0.2cm, 0.5cm, 0.1cm}, clip, width=\textwidth]{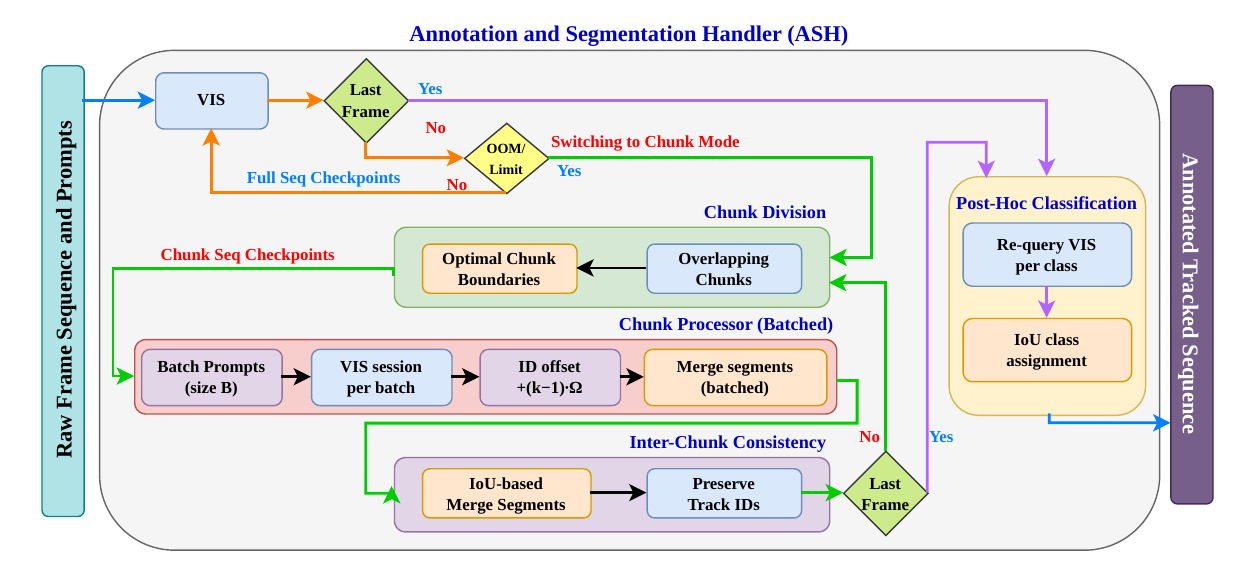}
\caption{Overview of the Annotation and Segmentation Handler (ASH) pipeline.
    ASH first attempts full-sequence Video Instance Segmentation (VIS) with
    chunk-level checkpointing. If an Out-of-Memory (OOM) error or length limit
    is encountered, the pipeline switches to chunk mode. The Chunk Division stage
    computes overlapping temporal segments and selects optimal chunk boundaries to
    maximise object density at transitions. Each chunk is then processed by the
    Chunk Processor, which partitions the full prompt set into batches of size $B$,
    runs an independent VIS session per batch, applies a per-batch Identity (ID)
    offset of $(k{-}1)\cdot\Omega$ to prevent collision across batches, and merges
    the resulting segments. The Inter-Chunk Consistency stage resolves local object
    IDs into globally consistent track IDs via overlap-region Intersection over
    Union (IoU) matching; if further chunks remain, the pipeline loops back with
    updated chunk sequence checkpoints. Once all chunks are processed, Post-Hoc
    Classification re-queries the VIS model independently for each class name in
    the dataset vocabulary and assigns semantic labels to tracked objects via IoU
    matching against per-class reference masks. The final annotated tracked
    sequence is written to the output.}
\label{fig2}
\end{figure}

\subsection{Annotation and Segmentation Handler (ASH)}

ASH implements an adaptive chunking strategy to handle video sequences of arbitrary length by dividing processing into manageable temporal segments while maintaining identity consistency across chunk boundaries.

The chunk-based processing subsystem consists of two primary components:

\begin{itemize}
    \item \textbf{Chunk Manager}: Coordinates the division of long sequences into overlapping chunks
    \item \textbf{Chunk Processor}: Handles individual chunk processing with boundary consistency
\end{itemize}

\textbf{Chunk Manager:}
The chunk manager employs a sliding window approach with configurable chunk size and overlap parameters. Given a video with frames indexed from $f_{\min}$ to $f_{\max}$, chunks are defined as:

\begin{equation}
\mathcal{C} = \{C_1, C_2, ..., C_n\} \text{ where } C_i = [s_i, e_i]
\end{equation}

\noindent where $\mathcal{C}$ represents the set of chunks and $C_i$ represents the $i$-th chunk spanning frames $s_i$ to $e_i$, with:

\begin{equation}
e_i = \min(s_i + \Delta - 1,\ f_{\max}), \quad s_{i+1} = e_i + 1
\end{equation}

\noindent where $\Delta$ is the configured chunk size. To improve identity matching at boundaries, the chunk manager identifies an optimal start frame within a search window around the nominal boundary by selecting the frame with the highest object density:

\begin{equation}
\text{OptimalStart}_i = \underset{f \in \mathcal{W}_i}{\arg\min}\ |f - s_i^{\text{nominal}}|
\quad \text{s.t.} \quad
|\mathcal{O}_f| = \max_{f' \in \mathcal{W}_i} |\mathcal{O}_{f'}|
\end{equation}

\noindent where $\mathcal{W}_i = [s_i - \delta,\ s_i + \delta]$ denotes the search window and $\mathcal{O}_f$ is the set of detected objects in frame $f$, $\delta$ defines the search window, and among frames tied for maximum object count, the one closest to the nominal boundary is selected.

\textbf{Chunk Processor:}
For each chunk $C_i$, the processor handles detection and segmentation through batched prompt processing. Rather than submitting all text prompts in a single forward pass, prompts are divided into batches of size $B$ and processed in separate Video Instance Segmentation (VIS) sessions:

\begin{equation}
\mathcal{P} = \{P_1, P_2, ..., P_m\} \text{ where } m = \lceil |\mathcal{P}| / B \rceil
\end{equation}

\noindent where $\mathcal{P}$ is the full set of prompts and each batch $P_k$ is processed in an independent VIS session. To prevent object ID collisions across batches within the same chunk, local IDs are shifted by a fixed offset per batch:

\begin{equation}
\text{UniqueID}_{k}(id) = id + (k-1) \cdot \Omega
\end{equation}

\noindent where $k$ is the batch index and $\Omega = 1000$ is the ID offset stride. Segments accumulated across all batches are then merged before the inter-chunk consistency step. This batched design allows ASH to handle datasets with large vocabulary sizes without exceeding GPU memory limits.

\textbf{Inter-Chunk Consistency:}
Identity continuity across chunks is achieved through overlap-region IoU matching. The final $\omega$ frames of chunk $C_i$ are retained and used in chunk $C_{i+1}$ to resolve local object IDs into consistent global IDs. For each object $O_j^B$ in the new chunk and each object $O_k^A$ from the previous chunk's overlap region, the average binary mask IoU over the overlap frames is computed:

\begin{equation}
\text{IoU}(O_j^B, O_k^A) = \frac{1}{|F_\text{overlap}|}\sum_{f \in F_\text{overlap}} \frac{|M(O_j^B, f) \cap M(O_k^A, f)|}{|M(O_j^B, f) \cup M(O_k^A, f)|}
\end{equation}

\noindent where $F_\text{overlap}$ is the set of overlapping frames and $M(O, f)$ is the binary mask of object $O$ at frame $f$. Global ID assignment is then performed as:

\begin{equation}
\text{GlobalID}(O_j^B) = \begin{cases} 
\text{ID}(O_k^A) & \text{if}\ \exists k : \text{IoU}(O_j^B, O_k^A) > \tau_{\text{overlap}} \\
              & \phantom{\text{if}\ } \text{and}\ \text{IoU}(O_j^B, O_k^A) \\
              & \phantom{\text{if}\ } = \max_k\ \text{IoU}(O_j^B, O_k^A) \\ 
\text{NewID}  & \text{otherwise}
\end{cases}
\end{equation}

\noindent where $\tau_{\text{overlap}}$ is the IoU threshold and unmatched objects are assigned a fresh global ID, ensuring newly appearing objects are correctly registered.

\textbf{Post-Hoc Class Determination:}
After tracking is complete, ASH performs a separate post-hoc classification pass to assign semantic class labels to each tracked object. For each globally identified object, a representative frame is selected at its first appearance. The underlying VIS model is then re-queried on that frame independently for each class name in the dataset vocabulary, generating reference masks per class. The tracked object's mask is matched against these reference masks via IoU:

\begin{equation}
\text{ClassID}(O_i) = \underset{c \in \mathcal{V}}{\arg\max}\ \text{IoU}(M(O_i, f_i^*),\ M_c^{\text{ref}})
\end{equation}

\noindent where $\mathcal{V}$ is the dataset vocabulary, $f_i^*$ is the representative frame for object $O_i$, and $M_c^{\text{ref}}$ is the reference mask generated by querying the VIS model with class name $c$. A minimum IoU threshold of $\tau_{\text{class}}$ is applied, below which the object is assigned an unknown class label. This decoupling of tracking and classification allows ASH to operate on open-vocabulary datasets without requiring class-specific training.

\textbf{Resume Support:}
ASH is designed with fault tolerance through chunk-level checkpointing. After each chunk is successfully processed, the current tracking state — including overlap segments, the next available global ID, and per-object detection scores — is persisted to disk. If processing is interrupted, ASH automatically resumes from the last completed chunk, making it practical for large-scale dataset annotation where processing times can span multiple sessions.
The overall integrated ASH pipeline architecture is drawn in Fig.~\ref{fig2}.

\section{Experiments}
\label{S:4}

\subsection{Datasets and Metrics}
\label{sub:DataMet}

We evaluate SAM3-ASH across two complementary dataset categories, each serving a distinct evaluation purpose. The first category comprises segmentation-annotated datasets---YouTube-VIS 2019/2021/2022 (validation sets) \cite{vis2019, vis2021, vis2022}; LV-VIS (validation and test sets) \cite{LVVIS}, an open-vocabulary benchmark covering a diverse label space beyond closed-set categories; OVIS (validation set) \cite{OVIS}; and MOTS20 (test set) \cite{MOTS}, which extends MOT17 with pixel-level segmentation annotations. Together, these benchmarks measure how faithfully the Annotation and Segmentation Handler (ASH) preserves the underlying VIS model's segmentation accuracy when processing is decomposed into temporal chunks. By directly comparing SAM3 against SAM3-ASH, we quantify the accuracy trade-off introduced by chunk-based processing relative to the original model's upper-bound performance. The second category comprises bounding-box-annotated datasets---MOT17 (test set) \cite{MOT17}, BDD100K (validation set) \cite{BDD100K}, DanceTrack (test set) \cite{DanceTrack}, and SportsMOT (test set) \cite{sportsmot}, a large-scale benchmark covering basketball, volleyball, and football scenes with fast-moving athletes and frequent occlusions. On these benchmarks, we compare SAM3-ASH against state-of-the-art MOT methods. To our knowledge, this is the first evaluation of a VIS-based tracker extended through ASH against dedicated MOT benchmarks, revealing the opportunities and challenges of applying memory-attention VIS models to long, unconstrained sequences. Notably, the juxtaposition of MOTS and MOT17 results provides additional insight into the impact of segmentation-based versus bounding-box-based ground truth on reported performance.

\noindent \textit{Metrics for Segmentation-Annotated Datasets:}
YouTube-VIS 2019/2021/2022 and OVIS are evaluated using mean Average Precision (\textbf{mAP}), which measures segmentation quality by averaging precision across IoU thresholds and object categories. LV-VIS additionally reports \textbf{mAP}$_{\text{base}}$ and \textbf{mAP}$_{\text{novel}}$, measuring performance separately on categories seen during training and unseen novel categories respectively, providing a direct assessment of open-vocabulary generalization. MOTS20 is evaluated using the Higher Order Tracking Accuracy (\textbf{HOTA}) \cite{HOTA} computed over pixel-level masks, which equally balances detection accuracy and association quality.

\noindent \textit{Metrics for Bounding-Box-Annotated Datasets:}
MOT17, BDD100K, DanceTrack, and SportsMOT are evaluated using three complementary metrics. \textbf{MOTA} \cite{MOTA} measures overall tracking performance by aggregating false positives, missed detections, and identity switches relative to ground truth, emphasizing detection quality. \textbf{IDF1} \cite{IDF1} focuses on identity preservation by measuring how consistently object identities are maintained across frames, prioritizing association quality over raw detection accuracy. \textbf{HOTA} \cite{HOTA} computed over bounding-box overlap provides a balanced assessment by equally weighting detection accuracy and association quality, addressing the limitations of both MOTA and IDF1 for long-term tracking evaluation.

\subsection{Implementation Details}
SAM3-ASH is evaluated on a single 40GB MIG partition of an NVIDIA H100 GPU,
with the full dataset vocabulary passed as the prompt set and prompts
processed in fixed batches of $B{=}5$ per VIS session. Video sequences are
divided into chunks of $\Delta{=}50$ frames with an overlap of
$\omega{=}10$ frames, the second reported configuration being
$\Delta{=}100$ with $\omega{=}40$; across all evaluated benchmarks peak GPU
memory consumption does not exceed 25\,GB. The remaining settings and the
rationale for this constrained memory budget are given in
Appendix~\ref{sub:ImpDet}.

\definecolor{oursgray}{rgb}{0.95,0.95,0.95}

\begin{table}[!tp]
\centering
\caption{Video Instance Segmentation results on \textbf{YouTube-VIS 2019, 2021,
and 2022 validation sets and the OVIS validation set}.
\textcolor{first}{\textbf{First place}},
\textcolor{second}{\textbf{Second place}},
\textcolor{third}{\textbf{Third place}}.
Formatting convention used throughout all tables.}
\label{tab:ytvis_ovis}
\setlength{\tabcolsep}{5pt}
\resizebox{\linewidth}{!}{%
\begin{tabular}{lllccccc}
\toprule
\multirow{2}{*}{\textbf{Method}} &
\multirow{2}{*}{\textbf{Pub.}} &
\multirow{2}{*}{\textbf{Backbone}} &
\multirow{2}{*}{\textbf{Training}} &
\multicolumn{3}{c}{\textbf{YouTube-VIS}} &
\multicolumn{1}{c}{\textbf{OVIS}} \\
\cmidrule(lr){5-7}\cmidrule(lr){8-8}
& & & &
\textbf{AP$_{19}$} & \textbf{AP$_{21}$} & \textbf{AP$_{22}$} &
\textbf{AP} \\
\midrule
OV2Seg~\cite{LVVIS}              & ICCV'23    & R50        & LVIS        & 27.2 & 23.6 & ---  & ---  \\
OVFormer~\cite{OVFormer}          & ECCV'24    & R50        & LV-VIS      & 34.8 & 29.8 & ---  & 15.1 \\
SOV~\cite{SOV}                    & TOMM'26    & R50        & LVIS        & 35.2 & 31.1 & ---  & ---  \\
BriVIS~\cite{BriVIS}              & AAAI'25    & R50        & LV-VIS      & 45.3 & 39.5 & ---  & 14.3 \\
GLEE-Lite~\cite{GLEE}             & CVPR'24    & R50        & Joint (5M+) & 53.1 & ---  & ---  & 27.1 \\
OV2Seg~\cite{LVVIS}              & ICCV'23    & SwinB      & LVIS        & 37.6 & 33.9 & ---  & 17.5 \\
OVFormer~\cite{OVFormer}          & ECCV'24    & SwinB      & LV-VIS      & 44.3 & 37.6 & ---  & 21.3 \\
EntitySAM~\cite{entitysam}        & CVPR'25    & ViT-S      & COCO        & 44.9 & 38.8 & ---  & 21.0 \\
SOV~\cite{SOV}                    & TOMM'26    & ConvNeXt-L & LVIS        & 50.9 & \textcolor{third}{\textbf{45.2}} & ---  & 23.1 \\
MASA~\cite{masa}                  & CVPR'24    & ViT-B      & SA-1B       & 51.8 & ---  & ---  & ---  \\
InstFormer~\cite{InstFormer}      & AAAI'25    & ViT-B/32   & YT-VIS      & 51.8 & ---  & ---  & ---  \\
GLEE-Plus~\cite{GLEE}             & CVPR'24    & Swin-L     & Joint (5M+) & \textcolor{third}{\textbf{63.6}} & ---  & ---  & 29.6 \\
OmniTracker-L$^\ddagger$~\cite{OmniTracker} & TPAMI'25 & Swin-L & Joint & \textcolor{second}{\textbf{63.9}} & --- & --- & --- \\
CPOVIS~\cite{CPOVIS}              & TPAMI'26   & ConvNeXt-L & Joint       & 62.6 & \textcolor{first}{\textbf{54.6}} & ---  & \textcolor{third}{\textbf{45.0}} \\
GLEE-Pro~\cite{GLEE}              & CVPR'24    & EVA02-L    & Joint (5M+) & \textcolor{first}{\textbf{67.4}} & ---  & ---  & 38.7 \\
SAM3~\cite{SAM3}                  & ICLR'26    & PE-L+      & SA-Co       & 57.4$^\dagger$ & ---  & ---  & \textcolor{first}{\textbf{60.5}} \\
\midrule
SAM3-ASH (Ours)                   & ---        & SAM3       & None        & 49.8 & 48.5 & 47.1 & 46.8 \\
\textbf{SAM3-ASH$^*$ (Ours)}      & ---        & SAM3       & None        &
\textbf{49.8} & \textcolor{second}{\textbf{48.5}} & \textcolor{first}{\textbf{47.1}} & \textcolor{second}{\textbf{48.7}} \\
\bottomrule
\end{tabular}}
\begin{tablenotes}\footnotesize
\item OV2Seg and SOV are trained on LVIS (image-only);
      InstFormer on YouTube-VIS (40 categories);
      EntitySAM on COCO without category annotations;
      GLEE variants jointly on 5M$+$ images.
      SAM3 uses SA-Co training data and prompts all 40 category names as text.
\item $^\ddagger$OmniTracker is jointly trained on COCO, MOT17, MOTS20,
      YTVIS19, and SOT/VOS datasets~\cite{OmniTracker} and does not operate
      in the open-vocabulary setting; it is included for reference only.
\item $^\dagger$SAM3 AP$_{19}$ is evaluated on the HQ-YTVIS test
      set~\cite{HQ-YTVIS}, a re-split of YTVIS 2019 training videos with
      refined boundary annotations; SAM3-ASH uses the standard YouTube-VIS
      2019 val set. The two AP$_{19}$ scores are not directly comparable.
\item CPOVIS reports open-vocabulary mAP on manually partitioned 2019/2021
      sets~\cite{CPOVIS}; 2022 and OVIS results are zero-shot.
      YT-VIS 2021/2022 results not reported by OmniTracker, SAM3, GLEE, or
      most prior methods.
\item $^*$SAM3-ASH on OVIS with $\Delta{=}100$, $\omega{=}40$:
      AP$_S$\,=\,28.4, HOTA\,=\,71.1, MOTA\,=\,65.2, IDF1\,=\,79.8.
      Standard setting ($\Delta{=}50$, $\omega{=}10$):
      AP$_S$\,=\,27.6, HOTA\,=\,70.6, MOTA\,=\,65.3, IDF1\,=\,79.1.
\end{tablenotes}
\end{table}

\begin{table}[!tp]
\centering
\caption{Open-vocabulary Video Instance Segmentation results on \textbf{LV-VIS}.
AP$_b$ and AP$_n$ denote Average Precision on base (641) and novel (555)
categories respectively.}
\label{tab:lvvis}
\setlength{\tabcolsep}{5pt}
\resizebox{\linewidth}{!}{%
\begin{tabular}{lllccccccc}
\toprule
\multirow{2}{*}{\textbf{Method}} &
\multirow{2}{*}{\textbf{Pub.}} &
\multirow{2}{*}{\textbf{Backbone}} &
\multirow{2}{*}{\textbf{Training}} &
\multicolumn{3}{c}{\textbf{Validation}} &
\multicolumn{3}{c}{\textbf{Test}} \\
\cmidrule(lr){5-7}\cmidrule(lr){8-10}
& & & & \textbf{AP} & \textbf{AP$_b$} & \textbf{AP$_n$} &
    \textbf{AP} & \textbf{AP$_b$} & \textbf{AP$_n$} \\
\midrule
InstFormer~\cite{InstFormer}     & AAAI'25   & ViT-B/32   & YT-VIS      & ---  & ---  & 12.2 & ---  & ---  & ---  \\
GLEE-Lite~\cite{GLEE}            & CVPR'24   & R50        & Joint (5M+) & 19.6 & 22.1 & 17.7 & ---  & ---  & ---  \\
OV2Seg~\cite{LVVIS}             & ICCV'23   & SwinB      & LVIS        & 21.1 & 27.5 & 16.3 & 16.4 & 23.3 & 11.5 \\
OVFormer~\cite{OVFormer}         & ECCV'24   & R50        & LV-VIS      & 21.9 & 22.1 & 21.8 & 15.2 & 18.0 & 13.1 \\
SOV~\cite{SOV}                   & TOMM'26   & R50        & LVIS        & 22.3 & 18.4 & 25.3 & 18.3 & 18.5 & 18.2 \\
GLEE-Pro~\cite{GLEE}             & CVPR'24   & EVA02-L    & Joint (5M+) & 23.9 & 24.6 & 23.3 & ---  & ---  & ---  \\
OVFormer~\cite{OVFormer}         & ECCV'24   & SwinB      & LV-VIS      & 24.7 & 26.8 & 23.1 & 19.5 & 23.1 & 16.7 \\
BriVIS~\cite{BriVIS}             & AAAI'25   & R50        & LV-VIS      & 27.7 & ---  & ---  & ---  & ---  & ---  \\
GLEE-Plus~\cite{GLEE}            & CVPR'24   & Swin-L     & Joint (5M+) & 30.3 & \textcolor{third}{\textbf{31.6}} & 29.3 & ---  & ---  & ---  \\
SOV~\cite{SOV}                   & TOMM'26   & ConvNeXt-L & LVIS        & \textcolor{third}{\textbf{38.0}} & 27.7 & \textcolor{second}{\textbf{45.9}} & \textcolor{third}{\textbf{33.0}} & \textcolor{third}{\textbf{26.0}} & \textcolor{second}{\textbf{38.3}} \\
SAM3~\cite{SAM3}                 & ICLR'26   & PE-L+      & SA-Co       & ---  & ---  & ---  & \textcolor{first}{\textbf{36.3}} & \textcolor{first}{\textbf{33.3}} & \textcolor{first}{\textbf{38.5}} \\
CPOVIS~\cite{CPOVIS}             & TPAMI'26  & ConvNeXt-L & Joint       & \textcolor{second}{\textbf{40.1}} & \textcolor{first}{\textbf{50.5}} & \textcolor{third}{\textbf{33.0}} & ---  & ---  & ---  \\
\midrule
\textbf{SAM3-ASH (Ours)}         & ---       & SAM3       & None        &
\textcolor{first}{\textbf{47.3}} & \textcolor{second}{\textbf{40.6}} & \textcolor{first}{\textbf{52.2}} & \textcolor{second}{\textbf{33.5}} & \textcolor{second}{\textbf{29.4}} & \textcolor{third}{\textbf{36.6}} \\
\bottomrule
\end{tabular}}
\begin{tablenotes}\footnotesize
\item Formatting as in Table~\ref{tab:ytvis_ovis}.
\item LV-VIS spans 1,196 categories: 641 base (AP$_b$) and 555 novel
      (AP$_n$).
      OVFormer is trained on LVIS then fine-tuned on LV-VIS video data.
      GLEE-Pro underperforms GLEE-Plus on LV-VIS due to sub-optimal
      resolution scaling~\cite{GLEE}.
\item AP$_b$/AP$_n$ breakdown not available for BriVIS (val only) and
      InstFormer (novel-class AP only).
      Test-set results not reported by GLEE, InstFormer,
      GLEE-Plus, or CPOVIS in their original papers.
      SAM3 reports test-set results only~\cite{SAM3}.
\end{tablenotes}
\end{table}

\begin{table}[!tp]
\centering
\caption{Multi-Object Tracking and Segmentation results on the \textbf{MOTS20}
\textbf{test set} (combined across all sequences).
Formatting as in Table~\ref{tab:ytvis_ovis}.}
\label{tab:mots}
\setlength{\tabcolsep}{3pt}
\begin{threeparttable}
\begin{tabular}{llccccc}
\toprule
\textbf{Method} & \textbf{Pub.} &
\textbf{HOTA} & \textbf{sMOTSA} & \textbf{DetA} &
\textbf{AssA} & \textbf{IDF1} \\
\midrule
TrackR-CNN~\cite{MOTS}               & CVPR'19  & ---  & 52.7 & ---  & ---  & ---  \\
TrackFormer~\cite{TrackFormer}        & CVPR'22  & ---  & 54.9 & ---  & ---  & 63.6 \\
OmniTracker-L$^\ddagger$~\cite{OmniTracker} & TPAMI'25 & --- & 67.5 & --- & --- & 69.2 \\
Seg2Track-SAM2$^\dagger$~\cite{Seg2Track-SAM2} & arXiv'25 & 61.1 & 55.9 & 57.9 & 66.0 & 72.2 \\
ReMOTSv2~\cite{ReMOTSv2}            & IVC'22   & \textcolor{third}{\textbf{65.4}} & 70.4 & 71.5 & 60.9 & 75.8 \\
EMNT~\cite{EMNT}                     & TIP'22   & \textcolor{second}{\textbf{66.0}} & 70.0 & 71.0 & 62.3 & 77.0 \\
\midrule
\textbf{SAM3-ASH$^\dagger$ (Ours)}  & ---      & \textcolor{first}{\textbf{68.1}} & 67.8 & 69.5 & 68.5 & 77.9 \\
\bottomrule
\end{tabular}
\begin{tablenotes}[flushleft]
\setlength{\labelsep}{0pt}
\footnotesize
\item HOTA jointly measures detection and association quality;
      sMOTSA is the segmentation-aware variant of MOTA.
      HOTA, DetA, and AssA for ReMOTSv2, EMNT, and Seg2Track-SAM2 are
      from the official MOTSChallenge leaderboard. TrackR-CNN,
      TrackFormer, and OmniTracker-L are not on the leaderboard;
      metrics sourced from their publications.
\item $^\dagger$Evaluated zero-shot, no dataset-specific training.
      SAM3 cannot process MOTS20 sequences due to memory constraints;
      SAM3-ASH is the only feasible evaluation path for this model family.
\item $^\ddagger$OmniTracker-L is jointly trained on COCO, MOT17, MOTS20,
      YTVIS19, and SOT/VOS datasets~\cite{OmniTracker}; all other methods
      use MOTS20-only training.
\end{tablenotes}
\end{threeparttable}
\end{table}

\begin{table}[!tp]
\centering
\caption{Results on the \textbf{MOT17} \textbf{test set} (COMBINED row,
averaged over the DPM, FRCNN, and SDP detector variants).
Formatting as in Table~\ref{tab:ytvis_ovis}.}
\label{tab:mot17}
\begin{threeparttable}
\setlength{\tabcolsep}{3pt}
\begin{tabular}{llcccc}
\toprule
\textbf{Method} & \textbf{Pub.}
& \textbf{HOTA}$\uparrow$ & \textbf{AssA}$\uparrow$
& \textbf{MOTA}$\uparrow$ & \textbf{IDF1}$\uparrow$ \\
\midrule
\multicolumn{6}{l}{\textit{Closed-vocabulary specialist trackers}} \\
BPMTrack~\cite{BPMTrack}              & TIP'24   & ---  & ---  & 81.3 & ---  \\
MSPNet~\cite{ZHENG2024110369}         & PR'24    & ---  & ---  & 74.6 & 71.9 \\
SambaMOTR~\cite{Samba}               & ICLR'25  & 59.7 & 59.7 & 72.9 & 71.0 \\
C-TWiX~\cite{CTWiX}                  & PR'25    & 63.1 & 62.5 & 78.1 & 76.3 \\
DiffMOT~\cite{DiffMOT}               & CVPR'24  & 64.5 & 64.6 & 79.8 & 79.3 \\
SparseTrack~\cite{Sparsetrack}       & CSVT'25  & 65.1 & ---  & 81.0 & 80.1 \\
NOOUGAT (Online)~\cite{NOOUGAT}      & IJCV'26  & 65.2 & 65.3 & 80.7 & 80.0 \\
SUSHI~\cite{SUSHI}                   & CVPR'23  & 66.5 & 67.8 & 81.1 & 83.1 \\
NOOUGAT (Offline)~\cite{NOOUGAT}     & IJCV'26  & \textcolor{third}{\textbf{66.9}} & 68.5 & 80.7 & 83.9 \\
CoNo-Link~\cite{CoNo-Link}           & AAAI'24  & \textcolor{second}{\textbf{67.1}} & 67.8 & 82.7 & 83.7 \\
TrackTrack~\cite{TrackTrack}         & CVPR'25  & \textcolor{second}{\textbf{67.1}} & 68.2 & 81.8 & 83.1 \\
SPAM~\cite{spam2024eccv}             & ECCV'24  & \textcolor{first}{\textbf{67.5}} & ---  & 80.7 & 84.6 \\
\midrule
\multicolumn{6}{l}{\textit{Unified trackers (closed-vocabulary)}} \\
OmniTracker-T$^\ddagger$~\cite{OmniTracker} & TPAMI'25 & 61.0 & ---  & 73.2 & 75.4 \\
OmniTracker-L$^\ddagger$~\cite{OmniTracker} & TPAMI'25 & 62.3 & ---  & 79.1 & 75.6 \\
\midrule
\multicolumn{6}{l}{\textit{Proposed (open-vocabulary, zero-shot)}} \\
\rowcolor{oursgray}
\textbf{SAM3-ASH$^\dagger$ (Ours)} & ---
    & \textbf{45.2} & \textbf{49.4}
    & \textbf{31.2} & \textbf{55.3} \\
\bottomrule
\end{tabular}
\begin{tablenotes}[flushleft]
\setlength{\labelsep}{0pt}
\footnotesize
\item The COMBINED row averages over DPM, FRCNN, and SDP detector
      variants; SDP-only scores are approximately 2--3 points higher.
      AssA not reported by SparseTrack, SPAM, or OmniTracker;
      HOTA and AssA not reported by BPMTrack or MSPNet.
      Additional SAM3-ASH metrics: DetA\,=\,42.4, LocA\,=\,75.7,
      IDSw\,=\,1577.
\item[$^\dagger$] Evaluated zero-shot via text prompts; no MOT17-specific
      training. Direct comparison with specialist trackers is indicative
      rather than competitive.
\item[$^\ddagger$] Jointly trained on COCO, MOT17, MOTS20, YTVIS19, and
      SOT/VOS datasets~\cite{OmniTracker}.
\end{tablenotes}
\end{threeparttable}
\end{table}

\begin{table}[!tp]
\centering
\caption{Comparison on the \textbf{BDD100K} \textbf{val set}.
Proposed and SAM3/SAM2MOT rows: masks converted to axis-aligned
bounding boxes, evaluated with CLEAR, HOTA, and TETA.
Closed-vocabulary baselines report published metrics.
Formatting as in Table~\ref{tab:ytvis_ovis}.}
\label{tab:bdd100k}
\begin{threeparttable}
\setlength{\tabcolsep}{3pt}
\begin{tabular}{llcccc}
\toprule
\textbf{Method} & \textbf{Pub.}
& \textbf{HOTA}$\uparrow$ & \textbf{MOTA}$\uparrow$
& \textbf{IDF1}$\uparrow$ & \textbf{TETA}$\uparrow$ \\
\midrule
\multicolumn{6}{l}{\textit{Closed-vocabulary (8-class)}} \\
UNINEXT~\cite{UNINEXT}        & CVPR'23  & ---  & 67.1 & 69.9 & ---  \\
SUSHI~\cite{SUSHI}            & CVPR'23  & ---  & 68.4 & 75.6 & ---  \\
SAM2MOT$^\dagger$~\cite{Sam2mot}  & AAAI'26 & --- & 57.5 & 70.8 & ---  \\
SAM2MOT$^\ddagger$~\cite{Sam2mot} & AAAI'26 & --- & 44.1 & 63.6 & ---  \\
GHOST~\cite{GHOSTTracker}     & CVPR'23  & \textcolor{second}{\textbf{61.7}} & 68.1 & 70.9 & ---  \\
BPMTrack~\cite{BPMTrack}      & TIP'24   & \textcolor{first}{\textbf{62.9}} & ---  & ---  & ---  \\
MASA~\cite{masa}              & CVPR'24  & ---  & ---  & 71.7 & \textcolor{second}{\textbf{54.2}} \\
\midrule
\multicolumn{6}{l}{\textit{Open-vocabulary (mask$\to$bbox)}} \\
GLEE (one NP)~\cite{GLEE}    & CVPR'24  & ---  & ---  & ---  & 14.9 \\
GLEE (all NPs)~\cite{GLEE}   & CVPR'24  & ---  & ---  & ---  & 18.0 \\
\midrule
\multicolumn{6}{l}{\textit{SAM3 ablations~\cite{SAM3} (mask$\to$bbox)}} \\
LLMDet + SAM3 tracker        & ICLR'26  & ---  & ---  & ---  & 28.9 \\
SAM3~\cite{SAM3}             & ICLR'26  & ---  & ---  & ---  & 47.2 \\
SAM3 + T-by-D tracker        & ICLR'26  & ---  & ---  & ---  & \textcolor{third}{\textbf{51.0}} \\
\midrule
\multicolumn{6}{l}{\textit{Proposed (mask$\to$bbox)}} \\
\rowcolor{oursgray}
\textbf{SAM3-ASH (Ours)}             & ---
    & \textcolor{third}{\textbf{56.9}} & \textbf{53.8}
    & \textbf{69.2} & \textcolor{first}{\textbf{66.2}} \\
\bottomrule
\end{tabular}
\begin{tablenotes}[flushleft]
\setlength{\labelsep}{0pt}
\footnotesize
\item[$^\dagger$] SAM2MOT with Co-DINO-l detector, 8-class, no fine-tuning.
\item[$^\ddagger$] SAM2MOT with Grounding-DINO-l detector, 8-class, no fine-tuning.
\item GLEE and SAM3 baselines report TETA only~\cite{SAM3};
      HOTA, MOTA, and IDF1 not published under this setting.
      Closed-vocabulary baselines report MOTA and IDF1 per their
      publications; TETA unavailable for those methods.
\end{tablenotes}
\end{threeparttable}
\end{table}

\begin{table}[!tp]
\centering
\caption{Comparison on the \textbf{DanceTrack} \textbf{test set}.
Methods grouped by paradigm.
Formatting as in Table~\ref{tab:ytvis_ovis}.}
\label{tab:dancetrack}
\begin{threeparttable}
\setlength{\tabcolsep}{3pt}
\begin{tabular}{llcccc}
\toprule
\textbf{Method} & \textbf{Pub.}
& \textbf{HOTA}$\uparrow$ & \textbf{AssA}$\uparrow$
& \textbf{MOTA}$\uparrow$ & \textbf{IDF1}$\uparrow$ \\
\midrule
\multicolumn{6}{l}{\textit{Tracking-by-detection}} \\
OC-SORT~\cite{OC-SORT}         & CVPR'23  & 54.6 & 40.2 & 89.6 & 54.6 \\
AED$^*$~\cite{AED}             & TIP'25   & 55.2 & 57.0 & 91.0 & ---  \\
SparseTrack~\cite{Sparsetrack} & CSVT'25  & 55.5 & 39.1 & 91.3 & 58.3 \\
C-TWiX~\cite{CTWiX}            & PR'25    & 62.1 & 47.2 & 91.4 & 63.6 \\
DiffMOT~\cite{DiffMOT}         & CVPR'24  & 62.3 & 47.2 & 92.8 & 63.0 \\
TrackTrack~\cite{TrackTrack}   & CVPR'25  & 66.5 & 52.9 & 93.6 & 67.8 \\
AED~\cite{AED}                 & TIP'25   & \textbf{66.6} & 69.7 & 92.2 & 69.7 \\
\midrule
\multicolumn{6}{l}{\textit{Graph-based (online/offline)}} \\
CoNo-Link~\cite{CoNo-Link}       & AAAI'24  & ---  & 50.7 & 89.7 & 64.1 \\
SUSHI~\cite{SUSHI}               & CVPR'23  & 63.3 & 50.1 & 88.7 & 63.4 \\
SPAM~\cite{spam2024eccv}         & ECCV'24  & 64.0 & ---  & 89.2 & 63.4 \\
NOOUGAT (Online)~\cite{NOOUGAT}  & IJCV'26  & 65.9 & 54.9 & 88.9 & 70.6 \\
NOOUGAT (Offline)~\cite{NOOUGAT} & IJCV'26  & \textcolor{third}{\textbf{68.4}} & 58.7 & 88.9 & 72.7 \\
MOTIP~\cite{MOTIP}               & CVPR'25  & \textcolor{second}{\textbf{73.7}} & 65.9 & 92.7 & 79.4 \\
\midrule
\multicolumn{6}{l}{\textit{Tracking-by-segmentation}} \\
SAM2MOT$^\dagger$~\cite{Sam2mot}  & AAAI'26 & \textcolor{first}{\textbf{75.5}} & 71.3 & 89.2 & 83.4 \\
SAM2MOT$^\ddagger$~\cite{Sam2mot} & AAAI'26 & \textcolor{first}{\textbf{75.8}} & 72.2 & 88.5 & 83.9 \\
\midrule
\multicolumn{6}{l}{\textit{Proposed (open-vocabulary, zero-shot)}} \\
\rowcolor{oursgray}
\textbf{SAM3-ASH (Ours)}         & ---      & \textbf{66.6} & 58.6 & 82.4 & 68.6 \\
\bottomrule
\end{tabular}
\begin{tablenotes}[flushleft]
\setlength{\labelsep}{0pt}
\footnotesize
\item[$^\dagger$] SAM2MOT with Co-DINO-l detector, no fine-tuning.
\item[$^\ddagger$] SAM2MOT with Grounding-DINO-l detector, no fine-tuning.
\item AssA not reported by SPAM, CoNo-Link, or NOOUGAT.
      MOTIP is trained directly on DanceTrack training data.
      AED$^*$ pretrained on TAO $\mathcal{C}^{\text{base}}$ only,
      without DanceTrack-specific training.
\end{tablenotes}
\end{threeparttable}
\end{table}

\subsection{Results and Discussion}
\label{sub:Results}

\textbf{Reading the Comparisons: Taxonomy and Regimes.}
Methods compared below fall into three categories.
\textit{Closed-vocabulary specialists} (e.g., ByteTrack~\cite{ByteTrack},
OC-SORT~\cite{OC-SORT}, DiffMOT~\cite{DiffMOT}, MOTIP~\cite{MOTIP}) are
trained on domain-specific data with a fixed category set, and their numbers
reflect optimization for the precise benchmark evaluated.
\textit{Trained open-vocabulary methods} (e.g., GLEE~\cite{GLEE},
CPOVIS~\cite{CPOVIS}, SOV~\cite{SOV}) extend to novel categories via text
prompts but are fully supervised on large curated video corpora.
\textit{SAM3-ASH} is training-free at the benchmark level, operating
zero-shot with no dataset-specific training, fine-tuning, or adaptation:
the underlying SAM3~\cite{SAM3} is trained on SA-Co, whose documented
source list contains none of the bounding-box MOT benchmarks evaluated here
(DanceTrack~\cite{DanceTrack}, MOT17~\cite{MOT17},
SportsMOT~\cite{sportsmot}), and while broad web sources cannot guarantee
zero frame-level incidental overlap, no structured annotation correspondence
exists --- the standard zero-shot claim in the field. Gaps to trained
methods should therefore be read not as deficits but as the cost of genuine
zero-shot generalization --- a cost that, as the results show, is often
surprisingly small.
A second distinction concerns the ground truth itself.
Segmentation-annotated benchmarks
(YouTube-VIS~\cite{vis2019,vis2021,vis2022}, LV-VIS~\cite{LVVIS},
OVIS~\cite{OVIS}, MOTS20~\cite{MOTS}) directly reward the mask fidelity and
identity consistency that memory-attention VIS models naturally produce,
whereas bounding-box-annotated benchmarks (MOT17~\cite{MOT17},
DanceTrack~\cite{DanceTrack}, BDD100K~\cite{BDD100K},
SportsMOT~\cite{sportsmot}) use boxes that enclose occluded objects in
full, while SAM3-ASH's boxes derive from masks of the \textit{visible}
region only. This structural mismatch suppresses IoU-based metrics on
box-annotated benchmarks independent of association quality, and should be
kept in mind when interpreting Part~II results.

\textbf{Part I: Segmentation-Annotated Benchmarks.}
Across YouTube-VIS 2019/2021/2022~\cite{vis2019,vis2021,vis2022} and
OVIS~\cite{OVIS} (Table~\ref{tab:ytvis_ovis}), SAM3-ASH achieves AP$_{19}$\,=\,49.8, AP$_{21}$\,=\,48.5,
AP$_{22}$\,=\,47.1, and OVIS AP up to 48.7 without any dataset-specific
training.
Results fall below SAM3's~\cite{SAM3} upper bound, as expected: chunk boundary
transitions introduce partial identity resets absent in full-sequence
processing.
Nevertheless, SAM3-ASH ranks second on YT-VIS 2021 and first on YT-VIS 2022
among open-vocabulary methods, and second on OVIS, confirming that the
degradation from chunking is bounded and tolerable in practice.

LV-VIS~\cite{LVVIS} provides the strongest evidence for SAM3-ASH's
open-vocabulary generalization (Table~\ref{tab:lvvis}).
Spanning 1,196 categories with 555 novel classes, it tests the regime where
SAM3's zero-shot grounding is most consequential.
SAM3-ASH achieves validation AP of 47.3 --- first overall --- with novel-class
AP of 52.2, outperforming all prior methods on this split.
Since SAM3~\cite{SAM3} reports test-set results only, the validation ranking
reflects SAM3's underlying capability with SAM3-ASH as its closest proxy.
On the held-out test set, where SAM3 is directly comparable, SAM3-ASH ranks
second (33.5~AP) behind SAM3 (36.3~AP), with the gap quantifying the cost of
chunk-based processing relative to full-sequence inference.
SAM3-ASH outperforms both GLEE~\cite{GLEE} and CPOVIS~\cite{CPOVIS} on
novel-class AP despite neither model having been trained on LV-VIS video data
--- though both have seen related video tracking distributions during training,
making this the most direct test of zero-shot open-vocabulary generalization
in our suite.

MOTS20~\cite{MOTS} is the most consequential result in this group
(Table~\ref{tab:mots}).
SAM3~\cite{SAM3} cannot be applied to MOTS20 sequences due to memory
constraints, making SAM3-ASH the only feasible evaluation path for this model
family on long sequences.
SAM3-ASH achieves HOTA\,=\,68.1 --- the highest reported score on this
benchmark --- surpassing ReMOTSv2~\cite{ReMOTSv2} (65.4) and
EMNT~\cite{EMNT} (66.0).
Both ReMOTSv2 and EMNT employ dataset-specific training and offline refinement;
SAM3-ASH uses neither, making its HOTA the strongest result reported under
genuinely zero-shot conditions on this benchmark.
The high AssA (68.5) confirms that identity consistency is maintained across
chunk boundaries, validating ASH's IoU-based inter-chunk matching.
This result demonstrates that ASH does not merely preserve base model quality
on sequences it can already handle --- it enables the model to reach sequences
entirely beyond its original scope, and does so competitively with specialized
supervised methods.

\textbf{Part II: Bounding-Box-Annotated Benchmarks.}
The structural mismatch described above is most directly quantified by the
MOTS20/MOT17 juxtaposition.
MOT17~\cite{MOT17} (Table~\ref{tab:mot17}) is the bounding-box-annotated
counterpart to the same pedestrian sequences underlying
MOTS20~\cite{MOTS}.
SAM3-ASH achieves HOTA\,=\,68.1 on MOTS20 and HOTA\,=\,45.2 on MOT17 --- a
23-point swing on identical video content, arising solely from annotation
format and IoU criterion.
SPAM~\cite{spam2024eccv}, the top-ranked specialist on MOT17, achieves
HOTA\,=\,67.5 there; SAM3-ASH's MOTS20 HOTA of 68.1 exceeds it.
This cross-benchmark comparison constitutes the clearest evidence that
box-annotated benchmarks structurally penalize mask-based models, and that
SAM3-ASH's association quality is competitive with the state of the art when
evaluated against annotations that match its output format.

On DanceTrack~\cite{DanceTrack} (Table~\ref{tab:dancetrack}),
SAM3-ASH achieves HOTA\,=\,66.6, matching
the best tracking-by-detection method (AED~\cite{AED}, TIP'25) despite
requiring no per-class detector training or domain-specific fine-tuning.
MOTIP~\cite{MOTIP}, the strongest overall entry at HOTA\,=\,73.7, is trained
directly on DanceTrack training data; the 7-point gap to SAM3-ASH should be
read in that context.
The practical upper bound for memory-attention segmentation models on this
benchmark is SAM2MOT~\cite{Sam2mot} (HOTA\,=\,75.8), which augments SAM2's
memory bank with a Cross-Object Interaction module that detects and purges
corrupted memory entries under pairwise occlusion.
SAM3~\cite{SAM3} inherits SAM2's memory bank without modification --- its
architectural advances (open-vocabulary zero-shot detector, masklet matcher)
do not address memory corruption under occlusion --- and SAM3-ASH shares this
limitation.
The gap to SAM2MOT therefore reflects the absence of explicit occlusion-memory
correction rather than weaker base tracking.
Notably, SAM2MOT does not report results on BDD100K~\cite{BDD100K}, suggesting
its Cross-Object Interaction does not generalize beyond the pairwise,
near-uniform-appearance occlusions that characterize DanceTrack.

On BDD100K~\cite{BDD100K} (Table~\ref{tab:bdd100k}), SAM3-ASH
outperforms SAM3~\cite{SAM3} directly,
ranking third on HOTA (56.9) and first on TETA among all reported methods.
The improvement over SAM3 arises from ASH's chunked temporal processing: by
not propagating memory of objects outside the current temporal window,
SAM3-ASH reduces erroneous cross-frame associations between visually similar
objects across categories --- a frequent failure mode in dense driving scenes.
SAM3's full-sequence memory bank accumulates identity confusion over entire
sequences; ASH's periodic resets act as implicit regularization on the memory
bank, yielding better inter-class discriminability on this benchmark.

On SportsMOT~\cite{sportsmot} (Table~\ref{tab:sportsmot}), SAM3-ASH
achieves HOTA\,=\,64.9 and
AssA\,=\,61.5.
HOTA is the most informative headline metric here, as it equally weights
detection and association quality without the detection bias that
afflicts MOTA~\cite{MOTA}.
The AssA of 61.5 exceeds ByteTrack~\cite{ByteTrack} trained on SportsMOT
(51.2) and is competitive with closed-vocabulary specialists, confirming strong
identity association despite the absence of domain-specific detection
fine-tuning.
SportsMOT's annotation policy assigns predicted boxes to occluded athletes only
when a visible body part remains, skipping those occluded beyond that
threshold.
This conditional policy narrows the bounding-box mismatch relative to
MOT17~\cite{MOT17}, since both the GT annotations and SAM3-ASH outputs
converge on the partial-visibility regime.
The residual HOTA gap to top methods is therefore attributable primarily to
weaker detection recall from the lack of domain-specific fine-tuning, and to
the same memory-bank occlusion sensitivity observed on DanceTrack, where
inter-athlete occlusion still produces GT boxes extending beyond the visible
mask region.

\textbf{Chunk Size Sensitivity.}
The accuracy trade-off introduced by ASH is governed by a single architectural
parameter: chunk size~$\Delta$.
As $\Delta$ increases toward the full sequence length, inter-chunk boundary
effects diminish and SAM3-ASH approaches SAM3's~\cite{SAM3} full-sequence
upper bound; as $\Delta$ decreases, the memory footprint shrinks at the cost
of more frequent identity resets.
This relationship holds across all benchmarks.
On OVIS~\cite{OVIS} --- the most occlusion-dense evaluation in our suite,
where temporal context is most valuable --- increasing $\Delta$ from 50 to 100
with overlap $\omega$ from 10 to 40 recovers AP from 46.8 to 48.7, closing
roughly half the gap to SAM3.
The OVIS result is representative: any dataset processed by SAM3-ASH can trade
memory budget for segmentation continuity by adjusting $\Delta$ and $\omega$,
with no retraining required.
Notably, peak GPU memory consumption remains below 25\,GB across all
benchmarks at the default $\Delta{=}50$, $\omega{=}10$ setting, leaving
substantial headroom to increase $\Delta$ on hardware where memory permits.

\begin{figure}[!t]
\centering
\includegraphics[width=\textwidth]{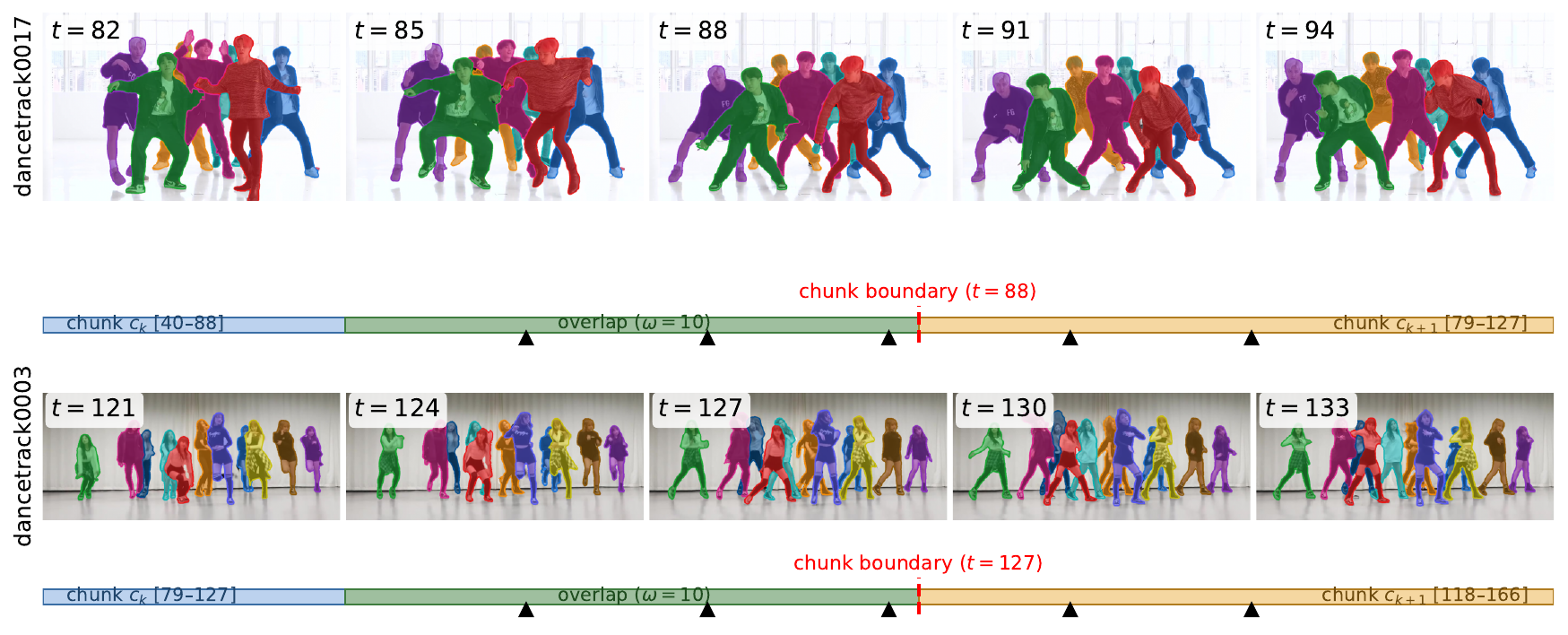}
\caption{Qualitative identity consistency across chunk boundaries
($\Delta{=}50$, $\omega{=}10$). Each row shows SAM3-ASH mask overlays
around an inter-chunk seam of a DanceTrack test sequence: frames left
of the dashed line are produced by chunk $c_k$, frames right of it by
chunk $c_{k+1}$, with colors keyed to the final merged identities
(cf.\ the IoU merge stage of Fig.~\ref{fig2}). Despite heavy
inter-dancer occlusion at both seams, every identity survives the
transition: overlap-window matching associates all objects one-to-one
(7/7 on \texttt{dancetrack0017}, 11/11 on \texttt{dancetrack0003}),
and the two chunks' masks agree at the seam frame itself with
0.91--0.995 IoU per object.}
\label{fig:qual_boundary}
\end{figure}

Fig.~\ref{fig:qual_boundary} makes this boundary behavior concrete: at
the first two chunk seams of two DanceTrack test sequences, IoU-based
matching over the $\omega{=}10$-frame overlap associates every active
track one-to-one across the boundary --- 18 of 18 objects over both
seams, with seam-frame mask agreement between the two chunks of
0.91--0.995 IoU --- while dancers occlude and cross through the seam
itself, the regime in which identity handover is hardest.

\subsection{Efficiency Analysis}
\label{sub:efficiency}

\noindent\textbf{Throughput Across Benchmarks.}
Table~\ref{tab:throughput} reports end-to-end inference throughput of
SAM3-ASH on a single 40\,GB MIG partition of an H100, measured over the
complete production runs behind the accuracy tables (chunked tracking
plus post-hoc classification; visualization and I/O excluded). Two
patterns stand out. First, throughput is governed by vocabulary size
rather than by scene complexity: single-class pedestrian and dancer
tracking sustains 2.1--2.5 FPS, while the same pipeline on the same
hardware drops monotonically as the prompt vocabulary grows --- 1.9 FPS
at 8 classes, 1.5 at 25, $\approx$0.7 at 40, and 0.017 at LV-VIS's
1{,}196 classes. This is exactly the $\lceil N/B \rceil$ scaling of
prompt-batch passes predicted by the GPT cost model with
$B{=}5$, and it quantifies why single-prompt inference is untenable for
open-vocabulary annotation. Second, absolute cost remains practical for
the annotation use case at realistic vocabulary sizes: annotating the
entire 39{,}973-frame BDD100K validation set over its 8-class
vocabulary takes under six GPU-hours.

\begin{table}[!tp]
\centering
\caption{SAM3-ASH inference throughput per benchmark on a single
40\,GB MIG H100 partition ($\Delta{=}50$, $\omega{=}10$, prompt batch
size $B{=}5$). Time covers chunked tracking and post-hoc
classification, excluding visualization and disk I/O.}
\label{tab:throughput}
\begin{threeparttable}
\setlength{\tabcolsep}{3pt}
\begin{tabular}{lrrrrr}
\toprule
\textbf{Benchmark} & \textbf{Vocab.\ ($N$)} & \textbf{Seq.} & \textbf{Frames}
& \textbf{Time (h)} & \textbf{FPS} \\
\midrule
DanceTrack test        & 1     & 35  & 38{,}551 & 4.3   & 2.51 \\
MOT17 test$^\dagger$   & 1     & 2   & 4{,}932  & 0.7   & 2.06 \\
BDD100K val            & 8     & 200 & 39{,}973 & 6.0   & 1.86 \\
OVIS val               & 25    & 140 & 8{,}784  & 1.6   & 1.51 \\
YT-VIS 2019 val        & 40    & 302 & 8{,}289  & 2.9   & 0.79 \\
YT-VIS 2021 val        & 40    & 421 & 13{,}195 & 5.0   & 0.74 \\
YT-VIS 2022 val        & 40    & 492 & 16{,}625 & 6.5   & 0.71 \\
LV-VIS val             & 1{,}196 & 837 & 19{,}139 & 317.2 & 0.017 \\
\bottomrule
\end{tabular}
\begin{tablenotes}[flushleft]
\setlength{\labelsep}{0pt}
\footnotesize
\item[$^\dagger$] Timing captured on sequences MOT17-01 and MOT17-06
      (each processed under the three detector copies of the public
      protocol); these are two of the four MOTS20 test sequences, so
      the figure transfers to MOTS20.
\end{tablenotes}
\end{threeparttable}
\end{table}

\noindent\textbf{Memory Behavior.}
Fig.~\ref{fig:memory_vs_frames} contrasts the memory trajectory of
full-sequence SAM3 with SAM3-ASH on the longest DanceTrack test
sequence. SAM3's memory bank grows with every processed frame: starting
from a 16.4\,GiB session footprint, peak allocation rises near-linearly
at $\approx$23\,MiB per frame and exhausts the 40\,GB partition at
frame 993 --- 62\% of the way through the sequence. SAM3-ASH's chunked
inference tears down each session after $\Delta{=}50$ frames, holding
the footprint to a flat $\approx$9.3\,GiB across all 41 chunks of the
same sequence on the same hardware. The plateau level is set by the
chunk length $\Delta$ and prompt batch size, not by the sequence, so
the gap between plateau and partition capacity is headroom that can be
spent on larger $\Delta$ or larger prompt batches rather than on
sequence length --- decoupling the hardware budget from video duration.

\noindent\textbf{Prompt Batching Throughput.}
To isolate the contribution of GPT's virtual prompt batching, we sweep
the prompt batch size $B$ over $\{1,2,5,10,20,40\}$ on a fixed
12-sequence subset of YT-VIS 2019 val (341 frames) with the full
40-class vocabulary on the same 40\,GB MIG partition. $B{=}1$
corresponds to sequential single-prompt inference --- the $O(N)$
baseline in which every prompt triggers its own image encoding and
propagation pass --- while $B{=}40$ processes all prompts in one
virtually batched pass. Fig.~\ref{fig:gpt_batching}(a) shows throughput
rising monotonically from 0.29 to 1.88 FPS, a $6.5\times$ speedup,
consistent with the $6.9\times$ full-vocabulary-versus-GT-prompt ratio
observed on matched production sequences. The measured curve is
reproduced within 5\% by a two-term cost model
$t_{\mathrm{frame}} = a\lceil N/B \rceil + c$, whose
batch-independent residual $c$ bounds the attainable speedup below the
ideal $N\times$: encoding cost is eliminated, propagation is not
(Appendix~\ref{app:batching}). 
Fig.~\ref{fig:gpt_batching}(b) reports the memory side of the
trade-off: peak reserved memory grows from 7.2\,GiB at $B{=}1$ to
26.9\,GiB at $B{=}40$, so even the fully batched setting fits the
40\,GB partition with headroom --- on this hardware the maximum
feasible batch is not smaller than the full YT-VIS vocabulary.

\noindent\textbf{Batching Equivalence.}
Virtual prompt batching is an inference-time restructuring and does not
alter model outputs: across 20 YT-VIS sequences, one batched
\texttt{add\_prompt} call and the equivalent single-prompt calls agree
exactly in detection count (agreement rate 1.0) with near-identical masks
(mean IoU 0.998). GPT's $O(1)$ image encoding is therefore obtained
without changing what the model predicts; the full protocol appears in
Appendix~\ref{app:batching}.

\subsection{Ablations and Generality}

The post-hoc classification threshold $\tau_{\mathrm{cls}}$ is the one
sensitive knob of the pipeline, while the chunking machinery itself proves
robust: the overlap $\omega$ shifts AP by at most 1.2 points across a
$20\times$ range and the inter-chunk matching threshold $\tau_{\mathrm{ov}}$
is inert across $[0.05, 0.5]$. Accuracy is thus governed by the
classification stage rather than by chunked inference itself; the full sweep
on OVIS val is reported in Appendix~\ref{sub:ablation}.

The claim that ASH is model-agnostic is substantiated empirically by a second
instantiation spanning a different backbone and a different prompt modality
--- box-prompted FLASH on SAM2~\cite{SAM2Auto}, prototyped in our earlier
SAM2Auto pipeline. The two instantiations sit within 4.6 HOTA of each other
on DanceTrack (62.0 vs.\ 66.6) and within 1.8 HOTA on MOT17 (43.4 vs.\ 45.2),
while overall performance tracks the quality of whatever detection front-end
feeds the handler --- the separation expected of a genuinely tracker-agnostic
layer. Appendix~\ref{sub:ash_sam2} reports the comparison in full.

\section{Path to Fully Automated Annotation}
\label{S:5}

SAM3-ASH's performance across segmentation-annotated and bounding-box-annotated
benchmarks reveals both the capabilities and the remaining limitations of
memory-attention segmentation trackers as a foundation for fully automated
tracking annotation. While SAM3-ASH demonstrates competitive zero-shot
performance across a wide range of sequences, three challenges must be
addressed before this paradigm can generalize to arbitrary video at scale.

\begin{enumerate}

    \item \textbf{Occlusion and Visual Similarity Handling.}
    The most consistent performance bottleneck across both benchmark categories
    is the inability of SAM2's memory bank to distinguish between visually
    similar objects under occlusion. As the DanceTrack and SportsMOT results
    demonstrate, memory-attention models accumulate identity confusion when
    objects of similar appearance interact, and no mechanism currently exists
    within SAM3 to detect and correct corrupted memory entries. Incorporating
    discriminative cues --- motion patterns, temporal consistency signals, or
    learned object-specific embeddings --- into the memory selection process
    would substantially improve identity preservation in crowded scenes and
    close the gap to methods such as SAM2MOT~\cite{Sam2mot} that address
    this explicitly.

    \item \textbf{Inter-chunk Identity Continuity in ASH.}
    ASH resolves the memory demand of full-sequence processing by decomposing
    inference into overlapping temporal chunks, enabling memory-attention
    segmentation trackers to operate on sequences of unbounded length with
    limited hardware resources. However, an object that is absent or occluded
    throughout the entire overlap window between two consecutive chunks will
    fail to match across the boundary and be assigned a new identity upon
    reappearance. This is a structural limitation of the chunking design
    independent of the base model's memory capacity, and it is the primary
    source of the performance gap between SAM3-ASH and SAM3 on long sequences.
    Extending the inter-chunk matching strategy --- for example, through
    appearance-based re-identification or longer overlap windows --- would
    narrow this gap and bring ASH closer to its base model's upper bound
    while preserving the memory efficiency that makes it practical.

    \item \textbf{Ground Truth Alignment for Segment-Based Trackers.}
    ASH unlocks the trackability of memory-attention segmentation models for
    unbounded sequences, pointing toward a new paradigm: fully automated
    tracking annotation of arbitrary video without domain-specific training.
    However, as the MOTS20/MOT17 juxtaposition demonstrates, existing
    bounding-box-annotated benchmarks do not accurately reflect the true
    capability of segment-based trackers. Current ground truths are designed
    to reward the complete predicted boxes that Kalman-filter-based trackers
    produce --- including predicted extents for occluded or partially visible
    objects --- systematically penalizing segment-based methods whose outputs
    faithfully enclose only the visible region. Updating ground truth
    annotations to reflect visible-region extents would align evaluation
    criteria with the output format of segmentation trackers and provide a
    more accurate measure of their true tracking capability.

\end{enumerate}

\noindent\textbf{Additional Limitations.}
Beyond these structural challenges, three failure modes merit note.
Very small or thin objects can fall below SAM3's detection threshold,
rendering them invisible to ASH regardless of chunking configuration.
Fast-moving objects that traverse the inter-chunk overlap window within
a few frames can defeat IoU-based matching, producing identity
fragmentation that longer overlaps only partially mitigate. Finally,
post-hoc class determination inherits the quality of its reference
masks: a corrupted mask in the representative frame can propagate an
incorrect label to an otherwise correct track.

Taken together, progress on memory-bank occlusion handling, inter-chunk
identity continuity, and ground truth alignment will determine how quickly
memory-attention segmentation trackers can realize their potential as the
backbone of fully automated, open-vocabulary tracking annotation pipelines
for video at scale.
On the engineering side, two directions are immediate: chunks are
mutually independent up to the identity-merge step, making multi-GPU
chunk parallelization a natural throughput multiplier, and hierarchical
vocabulary pruning would bound post-hoc classification cost on large
vocabularies such as LV-VIS's 1{,}196 classes. On the application side,
surveillance and egocentric video --- domains where per-domain
retraining is least practical --- are natural next deployments for a
text-promptable annotation engine.

\section{Conclusion}
\label{S:6}

We presented two contributions toward fully automated, open-vocabulary
tracking annotation of arbitrary video. The Generalized Presence Token
(GPT) extends SAM3's inference pipeline to handle multiple text prompts
simultaneously via a virtual prompt batching mechanism, reducing the
image encoding cost from $\mathcal{O}(N)$ to $\mathcal{O}(1)$ and
making open-vocabulary multi-object inference practical at scale. The
Annotation and Segmentation Handler (ASH) addresses the memory demand
of memory-attention segmentation trackers by decomposing full-sequence
inference into overlapping temporal chunks with IoU-based inter-chunk
identity matching, enabling SAM3 to process video sequences of unbounded
length under limited hardware resources without any retraining or
domain-specific adaptation.

Together, GPT and ASH form SAM3-ASH, a zero-shot open-vocabulary tracker
that operates via text prompts alone. Evaluated across eight benchmarks
spanning segmentation-annotated and bounding-box-annotated datasets,
SAM3-ASH achieves state-of-the-art HOTA on MOTS20 --- the only method
in its model family capable of processing these sequences at all ---
and remains competitive with trained specialists on MOT17, DanceTrack,
BDD100K, and SportsMOT without any dataset-specific training. The
MOTS20/MOT17 juxtaposition further establishes that bounding-box ground
truths systematically underestimate the association quality of
segment-based trackers, a finding with implications beyond this work.

The remaining gap to fully automated annotation is well-defined: occlusion
handling in the memory bank, inter-chunk identity continuity, and ground
truth alignment for segment-based evaluation. We view SAM3-ASH as a
practical baseline from which these challenges can be addressed, and
expect that resolving them will unlock memory-attention segmentation
trackers as the backbone of scalable, training-free video annotation
pipelines.

\section*{CRediT authorship contribution statement}
\textbf{Arash Rocky:} Conceptualization, Methodology, Software,
Validation, Investigation, Data curation, Visualization,
Writing -- original draft, review \& editing.
\textbf{Q.\,M. Jonathan Wu:} Supervision, Project administration,
Funding acquisition.

\section*{Declaration of competing interest}
The authors declare that they have no known competing financial
interests or personal relationships that could have appeared to
influence the work reported in this paper.

\section*{Declaration of generative AI and AI-assisted technologies
in the manuscript preparation process}
During the preparation of this work the authors used a generative AI
assistant (Claude, Anthropic) in order to assist with drafting and
editing portions of the text and with formatting. After using this
tool, the authors reviewed and edited the content as needed and take
full responsibility for the content of the published article.

\section*{Data availability}
All data used in this work are publicly available third-party
benchmarks: YouTube-VIS 2019/2021/2022~\cite{vis2019,vis2021,vis2022},
LV-VIS~\cite{LVVIS}, OVIS~\cite{OVIS}, MOTS20~\cite{MOTS},
MOT17~\cite{MOT17}, MOT20~\cite{MOT20}, DanceTrack~\cite{DanceTrack},
SportsMOT~\cite{sportsmot}, and BDD100K~\cite{BDD100K}, obtained from
their respective providers. No new datasets were created.

\section*{Acknowledgements}
\label{S:7}
This work was partially supported by TrustCAV, a CREATE program of
the Natural Sciences and Engineering Research Council of Canada
(NSERC). We also acknowledge the computational resources of the
Digital Research Alliance of Canada, provided through the NSERC
Alliance program, and thank the MOTChallenge team for access to
their evaluation server.

\clearpage

\clearpage

\appendix
\renewcommand{\thetable}{\Alph{section}.\arabic{table}}
\renewcommand{\thefigure}{\Alph{section}.\arabic{figure}}
\renewcommand{\theHtable}{\Alph{section}.\arabic{table}}
\renewcommand{\theHfigure}{\Alph{section}.\arabic{figure}}

\section{Implementation Details}
\label{sub:ImpDet}
SAM3-ASH is evaluated on a single 40GB MIG partition of an NVIDIA H100 GPU.
This conservative memory budget is intentional: by constraining inference to
a 40GB partition rather than the full GPU, we establish a reproducible baseline
that generalizes to a wide range of commonly available hardware configurations.
For each sequence, the full dataset vocabulary is used as the prompt set ---
all category names are passed as text prompts regardless of which classes are
present in the sequence, enabling open-vocabulary inference without
sequence-level supervision. For large-vocabulary datasets, prompts are
processed in fixed batches of $B{=}5$ per VIS session. Video sequences are
divided into chunks of $\Delta{=}50$ frames with an overlap of $\omega{=}10$
frames, and the ID offset stride is set to $\Omega{=}1000$ to prevent
cross-batch identity collisions. Inter-chunk identity matching uses an
IoU threshold of $\tau_{\text{overlap}}{=}0.2$, while post-hoc class
assignment applies a stricter minimum of $\tau_{\text{class}}{=}0.1$,
below which an object is labeled as unknown.
Across all evaluated benchmarks, peak GPU memory consumption does not
exceed 25\,GB, confirming that ASH provides a practical foundation for
extending any memory-attention segmentation tracker to video sequences
of unbounded length on broadly available hardware.

\setcounter{table}{0}
\setcounter{figure}{0}
\section{GPT Prompt Batching: Cost Model and Equivalence}
\label{app:batching}

\begin{figure}[t]
\centering
\includegraphics[width=\linewidth]{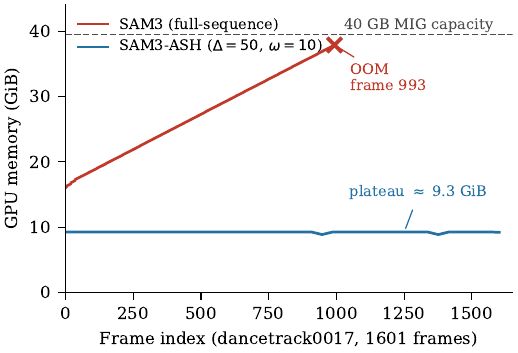}
\caption{GPU memory versus frame index on DanceTrack sequence
\texttt{dancetrack0017} (1{,}601 frames) on a 40\,GB MIG H100
partition. Full-sequence SAM3 accumulates memory-bank state roughly
linearly with sequence length and exhausts the partition at frame 993,
before completing the sequence. SAM3-ASH ($\Delta{=}50$, $\omega{=}10$)
bounds the per-chunk footprint to a constant $\approx$9.3\,GiB plateau,
independent of sequence length, completing all 1{,}601 frames.}
\label{fig:memory_vs_frames}
\end{figure}

\begin{figure}[t]
\centering
\includegraphics[width=\textwidth]{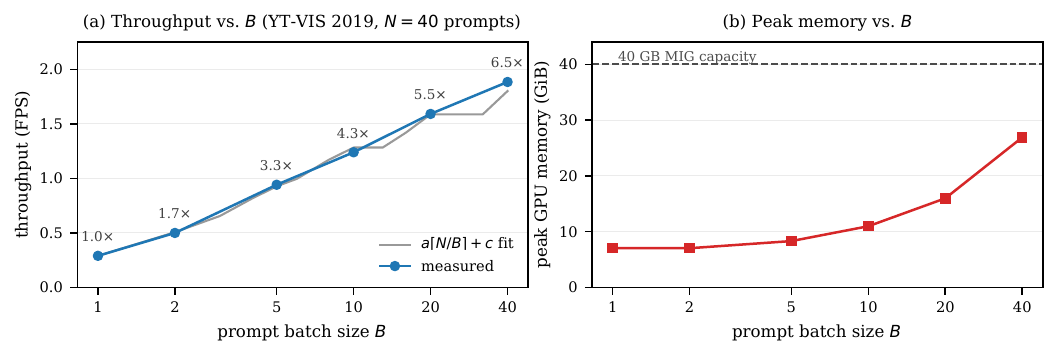}
\caption{GPT prompt-batching sweep on a fixed 12-sequence YT-VIS 2019
val subset (341 frames, full 40-class vocabulary, 40\,GB MIG H100).
(a) Inference-only throughput versus prompt batch size $B$, with
per-point speedups over the sequential $B{=}1$ baseline; the grey curve
is the two-term cost model $a\lceil N/B\rceil + c$. (b) Peak reserved
GPU memory versus $B$; all settings, including the fully batched
$B{=}40$, fit the 40\,GB partition.}
\label{fig:gpt_batching}
\end{figure}

This appendix expands the two batching results summarised in
Subsection~\ref{sub:efficiency}; the corresponding measurement plots
(Figs.~\ref{fig:memory_vs_frames} and~\ref{fig:gpt_batching}) are
reproduced here.

\noindent\textbf{Cost Model.}
The measured curve is explained by a two-term cost model, $t_{\mathrm{frame}} = a\lceil N/B
\rceil + c$ with $a{=}0.075$\,s per prompt pass and $c{=}0.48$\,s of
batch-independent work (propagation of active tracks, chunk merging,
classification), fitting all six points within 5\%. The residual term
$c$ bounds the attainable speedup below the ideal $N\times$: encoding
cost is eliminated, propagation is not.

\noindent\textbf{Batching Equivalence.}
Virtual prompt batching is an inference-time restructuring and must not
alter model outputs. We verify this at the frame level: for the first
frames of 20 YT-VIS sequences, we compare one batched
\texttt{add\_prompt} call over $N{=}10$ category prompts against ten
single-prompt calls (a one-element prompt list reduces to the original,
pre-GPT execution path), matching detections between the two runs by
greedy mask IoU since predictor outputs carry no per-prompt index.
Across the 10 frames with detections (28 matched objects), detection
counts agree exactly (agreement rate 1.0), matched masks are
near-identical (mean IoU 0.998, minimum 0.971), and confidence
deviations are bounded by $|\Delta p| \le 4.6\times10^{-3}$ (mean
$8.8\times10^{-4}$) --- the expected nondeterminism of bfloat16
autocast under reordered reductions rather than a semantic difference.
GPT's $O(1)$ image encoding is therefore obtained without changing what
the model predicts.

\setcounter{table}{0}
\section{Hyperparameter Sensitivity}
\label{sub:ablation}

Table~\ref{tab:ablation} varies each ASH hyperparameter around the
default configuration ($\Delta{=}50$, $\omega{=}10$,
$\tau_{\mathrm{ov}}{=}0.2$, $\tau_{\mathrm{cls}}{=}0.1$,
$\delta{=}\omega/2$) on OVIS val, using the full 25-class vocabulary
and the evaluation protocol of the main OVIS comparison
(Table~\ref{tab:ytvis_ovis}) throughout. The defaults were fixed a
priori and not tuned on this sweep.

The chunking machinery itself proves robust. The overlap length
$\omega$ shifts AP by at most 1.2 points across a $20\times$ range; the
inter-chunk matching threshold $\tau_{\mathrm{ov}}$ is inert across
$[0.05, 0.5]$ --- values of $0.05$ and $0.10$ yield byte-identical
predictions, as no inter-chunk match on OVIS val attains an overlap IoU
between the two --- and the boundary-search window $\delta$ changes AP
by at most 0.3. Enlarging the chunk to $\Delta{=}100$ (the paper's
second reported configuration) buys $+1.9$ AP at a higher per-chunk
memory plateau, making $\Delta$ the direct accuracy--memory dial of
Subsection~\ref{sub:efficiency}. One boundary of the design space is
worth stating: the adaptive boundary search requires
$\Delta - \omega - \delta \ge 1$ to guarantee forward progress; at
$\Delta{=}50$, $\omega{=}40$ this condition is violated and the
implementation falls back to a minimum-advance guard, so that setting
is reported for completeness rather than recommended.

The one sensitive knob is the post-hoc classification threshold
$\tau_{\mathrm{cls}}$, which degrades AP monotonically from 48.1 to
37.4 over $[0.05, 0.5]$. The sub-metrics expose the mechanism: stricter
thresholds discard uncertain classifications, so association quality
rises (AssA $82.6 \!\to\! 84.9$) while detection coverage collapses
(DetA $62.2 \!\to\! 49.3$). Accuracy is thus governed by the
classification stage rather than by chunked inference itself,
supporting the central claim that bounded-memory chunking does not
compromise tracking quality. Notably, $\tau_{\mathrm{cls}}{=}0.05$
already improves on the untuned default by $+1.3$ AP, indicating that
the reported numbers are a floor rather than a ceiling.

\begin{table}[!tp]
\centering
\caption{Sensitivity of SAM3-ASH to ASH hyperparameters on OVIS val
(full 25-class vocabulary). Each row varies one parameter from the
default configuration; the Defaults and $\Delta{=}100$ rows are the two
configurations reported in the main OVIS comparison.}
\label{tab:ablation}
\setlength{\tabcolsep}{3pt}
\footnotesize
\begin{threeparttable}
\begin{tabular}{llccccc}
\toprule
Axis & Value & AP & HOTA & DetA & AssA & IDF1 \\
\midrule
\multicolumn{2}{l}{Defaults ($\Delta{=}50$, $\omega{=}10$)} & 46.8 & 70.6 & 61.1 & 82.9 & 79.1 \\
\midrule
\multirow{4}{*}{$\omega$}
 & 2  & 46.2 & 70.0 & 60.6 & 82.4 & 78.5 \\
 & 5  & 45.6 & 69.9 & 60.6 & 82.2 & 78.3 \\
 & 20 & 46.9 & 70.4 & 60.7 & 83.1 & 79.1 \\
 & 40\tnote{$\dagger$} & 45.7 & 69.3 & 59.3 & 82.5 & 78.0 \\
\midrule
\multirow{3}{*}{$\tau_{\mathrm{ov}}$}
 & 0.05\,/\,0.10\tnote{$\ddagger$} & 46.8 & 70.6 & 61.1 & 82.9 & 79.1 \\
 & 0.35 & 46.8 & 70.6 & 61.1 & 82.9 & 79.1 \\
 & 0.50 & 46.5 & 70.4 & 61.0 & 82.7 & 78.8 \\
\midrule
\multirow{4}{*}{$\tau_{\mathrm{cls}}$}
 & 0.05 & 48.1 & 71.1 & 62.2 & 82.6 & 80.0 \\
 & 0.20 & 45.9 & 69.7 & 59.3 & 83.4 & 77.7 \\
 & 0.30 & 42.6 & 67.9 & 55.8 & 84.1 & 74.7 \\
 & 0.50 & 37.4 & 64.0 & 49.3 & 84.9 & 68.4 \\
\midrule
\multirow{3}{*}{$\delta$}
 & 1  & 46.7 & 70.5 & 61.0 & 82.9 & 79.1 \\
 & 2  & 46.5 & 70.4 & 60.9 & 82.8 & 79.0 \\
 & 10 & 46.5 & 70.5 & 61.2 & 82.7 & 78.9 \\
\midrule
$\Delta$ & 100 ($\omega{=}40$) & 48.7 & 71.1 & 61.3 & 83.8 & 79.8 \\
\bottomrule
\end{tabular}
\begin{tablenotes}[flushleft]
\setlength{\labelsep}{0pt}
\footnotesize
\item[$\dagger$] At $\Delta{=}50$, $\omega{=}40$ violates the
boundary-search validity condition $\Delta - \omega - \delta \ge 1$;
the run completes only under a forward-progress guard and is reported
for completeness.
\item[$\ddagger$] $\tau_{\mathrm{ov}}{=}0.05$ and $0.10$ produce
byte-identical predictions.
\end{tablenotes}
\end{threeparttable}
\end{table}

\setcounter{table}{0}
\section{ASH Beyond SAM3: A Second Instantiation on SAM2}
\label{sub:ash_sam2}

The claim that ASH is model-agnostic warrants empirical support beyond
a single backbone. ASH generalizes the chunked inference handler first
prototyped as FLASH in our earlier SAM2Auto pipeline~\cite{SAM2Auto},
whose architecture explicitly names an Annotation and Segmentation
Handler module driving SAM2~\cite{SAM2} with bounding-box prompts
supplied by an external open-vocabulary detector. The two
instantiations therefore differ along both axes relevant to the
agnosticism claim: the memory-attention backbone (SAM2 vs.\ SAM3) and
the prompt modality (boxes vs.\ text). Table~\ref{tab:ash_backbones}
juxtaposes the published FLASH results~\cite{SAM2Auto} with SAM3-ASH
on the three benchmarks common to both configurations.

\begin{table}[!tp]
\centering
\caption{ASH instantiated on two backbones with two prompt
modalities: box-prompted \textbf{FLASH} (ASH on
SAM2~\cite{SAM2Auto}) vs.\ text-prompted \textbf{SAM3-ASH}.
Bold marks the better score per benchmark.}
\label{tab:ash_backbones}
\begin{threeparttable}
\setlength{\tabcolsep}{3pt}
\begin{tabular}{llcccc}
\toprule
\textbf{Benchmark} & \textbf{Instantiation} & \textbf{Prompt}
& \textbf{HOTA}$\uparrow$ & \textbf{MOTA}$\uparrow$
& \textbf{IDF1}$\uparrow$ \\
\midrule
\multirow{2}{*}{DanceTrack}
& FLASH~\cite{SAM2Auto}    & box  & 62.0 & 64.1 & \textbf{72.5} \\
& \textbf{SAM3-ASH (Ours)} & text & \textbf{66.6} & \textbf{82.4} & 68.6 \\
\midrule
\multirow{2}{*}{MOT17$^\dagger$}
& FLASH~\cite{SAM2Auto}    & box  & 43.4 & 28.7 & \textbf{56.5} \\
& \textbf{SAM3-ASH (Ours)} & text & \textbf{45.2} & \textbf{31.2} & 55.3 \\
\midrule
\multirow{2}{*}{BDD100K}
& FLASH~\cite{SAM2Auto}    & box  & \textbf{58.8} & 11.1 & 55.8 \\
& \textbf{SAM3-ASH (Ours)} & text & 56.9 & \textbf{53.8} & \textbf{69.2} \\
\bottomrule
\end{tabular}
\begin{tablenotes}[flushleft]
\setlength{\labelsep}{0pt}
\footnotesize
\item FLASH additionally employs an online association module and
      checkpoint-based recovery absent from SAM3-ASH; both share the
      chunked ASH inference core.
\item[$^\dagger$] FLASH uses MOT17 private detections as box prompts;
      SAM3-ASH detects via text grounding, averaged over the
      triplicated public-server protocol (Table~\ref{tab:mot17}).
\end{tablenotes}
\end{threeparttable}
\end{table}

Three observations follow. First, on DanceTrack the two instantiations
sit within 4.6 HOTA of each other, and FLASH attains the highest IDF1
(72.5) reported in its comparison pool: long-term identity
preservation --- the property ASH is responsible for --- transfers
across backbones essentially intact. Second, on MOT17 the two
configurations are nearly indistinguishable (43.4 vs.\ 45.2 HOTA)
despite entirely different detection front-ends, and both trail box
specialists by a similar margin; the deficit is attributable to the
annotation-format mismatch quantified by the MOTS20/MOT17 juxtaposition
(Table~\ref{tab:mot17}) rather than to either backbone's tracking
layer. Third, BDD100K isolates the front-end: FLASH matches SAM3-ASH
on HOTA (58.8 vs.\ 56.9) but its MOTA collapses (11.1 vs.\ 53.8)
because ASH faithfully propagates the false positives injected by its
box-prompt detector, whereas SAM3's presence-token text grounding
suppresses them at the source --- with no change to ASH itself.

Across two backbones and two prompt modalities, the ASH layer thus
exhibits the same association behavior --- near-specialist identity
consistency with bounded chunk-boundary degradation --- while overall
performance tracks the quality of whatever detection front-end feeds
it. This separation is precisely the behavior expected of a genuinely
tracker-agnostic handler, substantiating the model-agnostic claim
empirically rather than by construction.

\setcounter{table}{0}
\setcounter{figure}{0}
\section{Additional Benchmark Comparison: SportsMOT}
\label{app:sportsmot}

Table~\ref{tab:sportsmot} reports the SportsMOT test-set comparison
discussed in Subsection~\ref{sub:Results}; it is placed here for
reasons of space only.

\begin{table}[!tp]
\centering
\caption{Comparison with state-of-the-art trackers on the
\textbf{SportsMOT} \textbf{test set}.
Methods are grouped by training data split.
Formatting as in Table~\ref{tab:ytvis_ovis}.}
\label{tab:sportsmot}
\begin{threeparttable}
\setlength{\tabcolsep}{2pt}
\begin{tabular}{llccccc}
\toprule
\textbf{Method} & \textbf{Pub.}
& \textbf{HOTA}$\uparrow$ & \textbf{DetA}$\uparrow$
& \textbf{AssA}$\uparrow$ & \textbf{MOTA}$\uparrow$
& \textbf{IDF1}$\uparrow$ \\
\midrule
\multicolumn{7}{l}{\textit{Closed-vocabulary, Train only}} \\
ByteTrack~\cite{ByteTrack}        & ECCV'22  & 62.8 & 69.8 & 51.2 & 94.1 & 77.1 \\
OC-SORT~\cite{OC-SORT}            & CVPR'23  & 71.9 & 72.2 & 59.8 & 94.5 & 86.4 \\
DiffMOT~\cite{DiffMOT}            & CVPR'24  & 72.1 & 72.8 & 60.5 & 94.5 & 86.0 \\
AED$^*$~\cite{AED}                & TIP'25   & 72.8 & 76.8 & 61.4 & 95.0 & 86.3 \\
Deep-EIoU~\cite{Deep-EIoU}        & WACV'24  & 74.1 & 75.0 & 63.1 & 95.1 & 87.2 \\
AED~\cite{AED}                    & TIP'25   & 77.0 & 80.0 & 68.1 & 95.1 & 87.1 \\
SambaMOTR$^\dagger$~\cite{Samba}  & ICLR'25  & 69.8 & 82.2 & 59.4 & 90.3 & 71.9 \\
\midrule
\multicolumn{7}{l}{\textit{Closed-vocabulary, Train+Val}} \\
ByteTrack~\cite{ByteTrack}        & ECCV'22  & 65.7 & 74.1 & 54.8 & 96.2 & 78.5 \\
MixSort-OC~\cite{sportsmot}       & ICCV'23  & 74.1 & 74.4 & 62.0 & 96.5 & 88.5 \\
DiffMOT~\cite{DiffMOT}            & CVPR'24  & 76.2 & 76.1 & 65.1 & 97.1 & 89.3 \\
Deep-EIoU~\cite{Deep-EIoU}        & WACV'24  & 77.2 & 79.8 & 67.7 & 96.3 & 88.2 \\
AED~\cite{AED}                    & TIP'25   & \textcolor{third}{\textbf{79.1}} & 81.8 & 70.1 & 97.1 & 89.4 \\
\midrule
\multicolumn{7}{l}{\textit{Graph-based (online/offline)}} \\
NOOUGAT (Online)~\cite{NOOUGAT}   & IJCV'26  & \textcolor{second}{\textbf{81.0}} & ---  & 74.3 & 96.0 & 85.3 \\
NOOUGAT (Offline)~\cite{NOOUGAT}  & IJCV'26  & \textcolor{first}{\textbf{85.6}} & ---  & 83.0 & 95.9 & 92.3 \\
\midrule
\multicolumn{7}{l}{\textit{Proposed (open-vocabulary, zero-shot)}} \\
\rowcolor{oursgray}
\textbf{SAM3-ASH (Ours)}          & ---      & \textbf{64.9} & 68.9 & \textbf{61.5} & 75.8 & 70.8 \\
\bottomrule
\end{tabular}
\begin{tablenotes}[flushleft]
\setlength{\labelsep}{0pt}
\footnotesize
\item All closed-vocabulary methods use YOLOX as the detector.
\item[$^*$] AED$^*$ uses TAO $\mathcal{C}^{\text{base}}$ pretraining with
      YOLOX trained on SportsMOT; all other Train-only methods use YOLOX
      trained on SportsMOT only.
\item[$^\dagger$] SambaMOTR training split not explicitly specified
      in the original paper~\cite{Samba}; placed in Train-only as the
      conservative assumption.
\item DetA not reported by NOOUGAT.
      SAM3-ASH: evaluated zero-shot, no dataset-specific training.
\end{tablenotes}
\end{threeparttable}
\end{table}

\end{document}